\documentclass[runningheads]{llncs}

\usepackage{eccv}

\usepackage{eccvabbrv}

\usepackage{graphicx}

\usepackage{amsmath}
\usepackage{amssymb}
\usepackage{booktabs}
\usepackage{multirow}
\usepackage{xcolor}
\usepackage{algorithm}
\usepackage{algorithmic}
\usepackage{subcaption}
\usepackage{enumitem}
\usepackage{microtype}
\usepackage{xspace}
\usepackage{bm}
\usepackage{pifont}

\newcommand{\simclr}{SimCLR\xspace}
\newcommand{\vicreg}{VICReg\xspace}
\newcommand{\barlowtwins}{Barlow Twins\xspace}
\newcommand{\byol}{BYOL\xspace}
\newcommand{\dino}{DINO\xspace}
\newcommand{\diet}{DIET\xspace}

\usepackage[accsupp]{axessibility}  

\usepackage{hyperref}

\usepackage{orcidlink}

\begin{document}

\title{IConE: Batch Independent Collapse Prevention for Self-Supervised Representation Learning} 

\titlerunning{IConE: Small Batch Representation Learning}

\author{Konstantinos Almpanakis\orcidlink{0009-0005-3361-9488} \and
Anna Kreshuk\orcidlink{0000-0003-1334-6388}}

\authorrunning{K.~Almpanakis and A.~Kreshuk}

\institute{European Molecular Biology Laboratory, Heidelberg, Germany\\
\email{\{konstantinos.almpanakis,anna.kreshuk\}@embl.de}}

\maketitle

\begin{abstract}
    Self-supervised learning (SSL) has revolutionized representation learning, with Joint-Embedding Architectures (JEAs) emerging as an effective approach for capturing semantic features. Existing JEAs rely on implicit or explicit batch interaction -- via negative sampling or statistical regularization -- to prevent representation collapse. This reliance becomes problematic in regimes where batch sizes must be small, such as high-dimensional scientific data, where memory constraints and class imbalance make large, well-balanced batches infeasible. 
    We introduce \textbf{IConE} (\textbf{I}nstance-\textbf{Con}trasted \textbf{E}mbeddings), a framework that decouples collapse prevention from the training batch size. Rather than enforcing diversity through batch statistics, IConE maintains a global set of learnable auxiliary instance embeddings regularized by an explicit diversity objective. This transfers the anti-collapse mechanism from the transient batch to a dataset-level embedding space, allowing stable training even when batch statistics are unreliable, down to batch size 1. Across diverse 2D and 3D biomedical modalities, IConE outperforms strong contrastive and non-contrastive baselines throughout the small-batch regime (from $B=1$ to $B=64$) and demonstrates marked robustness to severe class imbalance. Geometric analysis shows that IConE preserves high intrinsic dimensionality in the learned representations, preventing the collapse observed in existing JEAs as batch sizes shrink.


\end{abstract}
\section{Introduction}
\label{sec:introduction}

The success of Self-Supervised Learning (SSL) has led to powerful general-purpose vision models, such as DINOv2~\cite{oquab2023dinov2} and its successor DINOv3~\cite{simeoni2025dinov3}, which were transformative for downstream tasks across diverse domains~\cite{brondolo2025dinov2,ayzenberg2024dinov2,neuschmied2026improving,moutakanni2025cell,chen2026driving}. A major line of work in SSL is based on reconstruction, where representations are learned by recovering input content from a corrupted observation~\cite{he2022masked}. In contrast, multi-view or joint-embedding approaches learn by enforcing consistency between multiple augmented views of the same instance while preventing representational collapse~\cite{chen2020simple, grill2020bootstrap, bardes2021vicreg, caron2021emerging}. Empirically, joint-embedding methods have proven highly effective for learning discriminative representations in vision~\cite{van2025joint}.

Unlike natural 2D images, most scientific domains lack foundation models as the Internet-scale paradigm~\cite{oquab2023dinov2} does not readily apply. With new specialized modalities continuously emerging and no ImageNet~\cite{deng2009imagenet} equivalent for 3D medical volumes or hyperspectral data, waiting for massive datasets is impractical. Consequently, self-supervised learning must be effective not just at scale, but in the small-data regimes typical of scientific discovery.

These regimes impose not only data scarcity but also strict computational constraints. In many scientific applications (\eg volumetric CT/MRI, remote sensing, materials science), a single instance -- such as a 3D volume or video sequence -- can saturate GPU memory, limiting training to batch sizes of $1$-$2$ samples. Conversely, increasing batch size to match standard SSL practice (\eg $B>256$) becomes problematic when datasets are small (\eg $N<5000$), as large batches relative to dataset size reduce gradient stochasticity and weaken the implicit regularization relied upon by many objectives~\cite{keskar2016large}. Severe class imbalance further exacerbates small-batch instability, as individual batches may fail to represent minority classes. As a result, existing SSL methods often degrade in small-batch, small-data regimes~\cite{assran2022hidden,chen2022we}.

Theoretical analysis~\cite{wang2022chaos} provides further motivation for our work:  contrastive SSL can be understood as aligning augmentation manifolds. In higher-dimensional spaces, these manifolds are inherently sparse, requiring stronger augmentations to induce the semantic overlap for representation learning. This demands stronger regularization to prevent collapse, creating a paradox: high-dimensional data requires robust regularization, but its massive memory footprint dictates small batches, precisely where batch-dependent regularization becomes unreliable.

Current joint-embedding methods balance two coupled objectives: multi-view consistency and collapse prevention. In standard methods, both are implemented within the encoder’s representation space and rely on batch-level interactions during the forward pass. This tight coupling makes collapse prevention dependent on reliable batch statistics, which deteriorate as batch size shrinks. In contrast, we show that decoupling these objectives enables robust optimization independent of batch size. Conceptually, this shifts collapse prevention from a stochastic batch-level interaction to a deterministic dataset-level geometric constraint.\newline



\begin{figure*}[t]
    \centering
    \includegraphics[width=\textwidth]{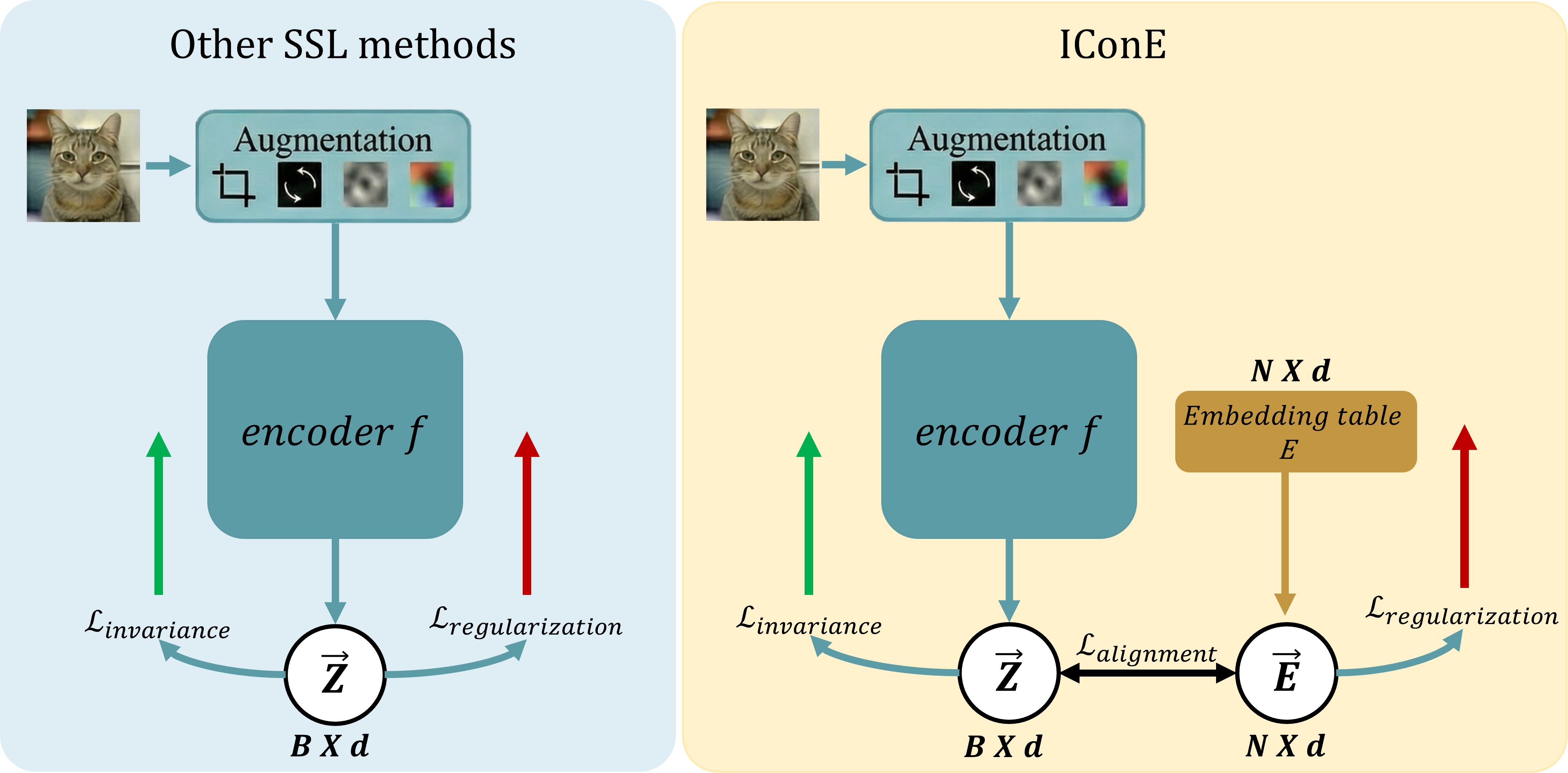}
    \caption{\textbf{Decoupling invariance and anti-collapse gradients with IConE.} \textbf{Left:} Standard JEAs apply both the multi-view invariance objective ($\mathcal{L}_{\text{invariance}}$) and the anti-collapse mechanism ($\mathcal{L}_{\text{regularization}}$) directly within the encoder's dynamic representation space $\vec{Z}$. This couples the anti-collapse constraint to the current batch. \textbf{Right:} IConE transfers the anti-collapse mechanism to an explicitly different parameter space. A persistent embedding table yields instance-specific anchors $\vec{E}$, which are independently structured via $\mathcal{L}_{\text{regularization}}$. The encoder $f$ is optimized purely through attractive forces: local view consistency ($\mathcal{L}_{\text{invariance}}$) and alignment to its regularized target ($\mathcal{L}_{\text{alignment}}$). This structural decoupling removes the need for batch-dependent negative sampling or variance statistics, applying the repulsive force to the entire dataset at the same time.}
    \label{fig:IConE_vs_others}
\end{figure*}

\noindent\textbf{IConE.}
We propose \textbf{IConE} (\textbf{I}nstance-\textbf{Con}trasted \textbf{E}mbeddings), a framework designed to decouple the multi-view learning process from the anti-collapse mechanism (see ~\cref{fig:IConE_vs_others}) by introducing an auxiliary embedding for each training instance that provides repulsive regularization. These embeddings are produced by a learnable embedding table indexed by instance IDs, with regularization applied directly in this parameter space across the entire dataset rather than only within the current batch. The auxiliary embeddings are used only during training; at inference time the encoder operates as a standard feature extractor without requiring instance identifiers. The main encoder learns consistency under augmentation, but also aligns augmented instance views with these regularized targets via cosine similarity maximization. By applying a diversity penalty to the auxiliary embedding table, we ensure they remain well-separated on the hypersphere, providing the geometric structure needed to prevent collapse without relying on batch composition. IConE therefore defines a repulsive objective that is invariant to the batch size, allowing the optimization to proceed stably regardless of memory constraints.\newline


\noindent{\textbf{Contributions.}}
\begin{itemize}[leftmargin=*,itemsep=2pt]
    \item We propose \textbf{IConE}, a minimal SSL method that explicitly separates multi-view invariance learning and the anti-collapse mechanism needed for joint-embedding architectures. Across diverse 2D and 3D modalities, IConE achieves state-of-the-art linear-probe performance in small-batch ($1$-$64$) and low-data regimes.
    \item Beyond standard downstream image classification, we provide a comprehensive geometric analysis of the learned representations. We show that IConE achieves higher rank and intrinsic dimensionality than baselines, indicating reduced dimensional collapse even when training on single-instance batches.
    \item We establish the robustness of IConE on highly unbalanced datasets, demonstrating that the instance embedding table provides a global regularization signal that prevents the poor feature learning for minority classes typically observed in batch-dependent methods.
    \item We extend the core principle of IConE to supervised representation learning. By applying the same parametric anchoring mechanism guided by class labels, we overcome the batch dependency of Supervised Contrastive Learning (SupCon) and demonstrate superior performance over both SupCon and standard Cross-Entropy baselines in constrained regimes.
\end{itemize}

\section{Related Work}
\label{sec:related}

Modern Joint-Embedding Architectures (JEAs) learn representations by enforcing instance-level consistency across augmented views while preventing trivial outputs~\cite{wang2020understanding,wang2022chaos}. To avoid representational collapse, existing methods utilize either explicit instance discrimination or implicit architectural and statistical constraints. Crucially, both paradigms rely on the training batch to implement the anti-collapse mechanisms. While highly effective at scale, this fundamental batch dependency creates the exact bottleneck that IConE is designed to resolve. We categorize existing approaches based on their mechanism for collapse prevention and their resulting dependency on batch size.

\subsection{Batch-Dependent Representation Learning}

\noindent\textbf{Explicit Interaction (Contrastive Learning).}
Methods like \simclr~\cite{chen2020simple} and others\cite{oord2018representation,he2020momentum} minimize the InfoNCE loss by contrasting positive pairs against in-batch negative samples. This creates a strict dependency on large batches to adequately approximate the negative data distribution. At extremely small batch sizes, this limited negative pool causes learning to stall~\cite{chen2022we}. Furthermore, under significant class imbalance, minority classes are often entirely absent from small batches, depriving the model of the gradients needed to distinguish them and resulting in poor minority class representations~\cite{jiang2021self}.\newline

\noindent\textbf{Implicit Interaction (Non-Contrastive Learning).}
To avoid explicit negatives, methods like \vicreg~\cite{bardes2021vicreg}, \barlowtwins~\cite{zbontar2021barlow}, and W-MSE~\cite{ermolov2021whitening} regularize statistical properties of batch embeddings. However, in very small batches, sample variance becomes ill-conditioned, rendering this anti-collapse mechanism unreliable. Asymmetric distillation approaches similarly rely on batch dynamics: \dino~\cite{caron2021emerging} centers representations using an exponential moving average of the batch mean, which becomes destabilized by individual samples in small batches. Likewise, while \byol~\cite{grill2020bootstrap} can theoretically function without Batch Normalization~\cite{richemond2020byol}, standard implementations exhibit lower performance as batch size decreases. Ultimately, because batch statistics are inherently unreliable in the small-batch limit, these methods require significant modification to function effectively~\cite{ioffe2017batch,buglakova2025tiling}.
Methods like SwAV~\cite{caron2020unsupervised} learn a set of prototypes and assign samples to them. While SwAV also learns embedding vectors (prototypes), these are shared across instances ($K \ll N$), requiring an assignment mechanism that depends on batch statistics. This step is computed over the batch using the Sinkhorn-Knopp algorithm; consequently, SwAV requires sufficiently large batches to approximate the dataset distribution and perform meaningful clustering assignments.
\subsection{Approaches for Reducing Batch Size Dependency}

Several approaches aim to reduce the impact of smaller batches on downstream performance. MoCo~\cite{he2020momentum,chen2020improved,chen2021empirical} decouples negative sampling from batch size via a feature queue. However, ensuring consistency among these past encoded features requires a momentum encoder, doubling model parameters. Furthermore, queue dynamics assume older embeddings remain semantically relevant, which often fails during early training when the encoder evolves rapidly. IConE sidesteps these issues entirely: instead of caching potentially stale outputs, we optimize persistent instance embeddings directly via gradient descent, ensuring they remain synchronized with the evolving encoder.
Beyond memory banks, several other works have recognized the batch-size sensitivity of SSL and proposed mitigation strategies, including loss decoupling, per-instance moving averages, or multi-view generation~\cite{yeh2022decoupled,yuan2022provable,zhang2023pretext,kuang2025piccl,pototzky2022fastsiam}. However, these approaches largely retain momentum encoders, memory banks, or in-batch negatives, and thus do not fully eliminate batch dependency. Moreover, they typically define small batch as 64--256 samples (infeasible for high-dimensional volumes) and evaluate on large-scale natural image benchmarks rather than data-scarce scientific regimes. To our knowledge, IConE is the first method to demonstrate robust performance in extremely small batch settings, providing comprehensive evaluation across 2D and 3D modalities for batch sizes from 1 to 64.\newline

\noindent\textbf{Parametric Instance Discrimination.}
Our work revisits Parametric Instance Discrimination (PID)~\cite{wu2018unsupervised}, which treats each instance as a unique class. Recent work like \diet~\cite{balestriero2023unsupervised} has shown that PID can be competitive with modern SSL if tuned correctly. However, \diet's cross-entropy framework introduces two structural limitations: it enforces multi-view alignment only implicitly through shared instance labels, and it separates representations via implicit softmax competition rather than direct geometric constraints. IConE addresses these issues by augmenting instance discrimination with an explicit view-view consistency term, and by replacing softmax competition with a direct geometric regularizer on the auxiliary embedding space.
\section{Method}
\label{sec:method}

 Our framework is built on the insight that the geometric requirements of representation learning---uniformity and alignment~\cite{wang2020understanding}---can be decoupled into separate optimization processes. Alignment is learned in the encoder representation space, while uniformity is enforced in a persistent auxiliary embedding space. By offloading the uniformity constraint to a set of persistent instance embeddings, we free the encoder from the need to observe a representative set of negative samples within a single batch.

\subsection{Preliminaries}
\label{sec:method:prelim}

Let $\mathcal{D} = \{\mathbf{x}_1, \ldots, \mathbf{x}_N\}$ be a dataset of $N$ unlabeled samples. We aim to learn an encoder $f_\theta: \mathcal{X} \to \mathbb{R}^d$ that maps high-dimensional inputs to discriminative representations. For a given sample $\mathbf{x}_i$, we generate $V$ augmented views $\{\mathbf{x}_i^{(v)}\}_{v=1}^V$ by sampling from a stochastic augmentation family $\mathcal{T}$. The encoder outputs are projected onto the unit hypersphere $\mathbb{S}^{d-1} = \{ \mathbf{z} \in \mathbb{R}^d : \|\mathbf{z}\|_2 = 1 \}$:
\begin{equation}
    \mathbf{z}_i^{(v)} = \frac{f_\theta(\mathbf{x}_i^{(v)})}{\|f_\theta(\mathbf{x}_i^{(v)})\|_2}.
\end{equation}
Unlike methods that employ a projector network between the backbone and the loss \cite{bardes2021vicreg,chen2020simple}, IConE operates directly on backbone representations. This eliminates the ambiguity of choosing between projected and backbone features at evaluation time.
In this work, $B$ refers to the batch size of unique instances loaded from the dataset, resulting in $B \times V$ total forward passes per step.

\subsection{Instance Embeddings as Geometric Anchors}
\label{sec:method:embeddings}

To remove the batch-size dependency, we define the target geometry of the latent space explicitly using an auxiliary embedding set. We maintain a learnable embedding table $\mathbf{E} \in \mathbb{R}^{N \times d}$, where the row $\mathbf{e}_i$ serves as a persistent anchor for instance $i$.
These embeddings are initialized from a normal distribution $\mathcal{N}$. During training, we normalize embeddings when computing similarities: $\tilde{\mathbf{e}}_i = \mathbf{e}_i / \|\mathbf{e}_i\|_2$. The unnormalized parameters $\mathbf{e}_i$ are updated directly by gradient descent. This table allows us to separate the optimization into two distinct objectives:
\begin{enumerate}
    \item \textbf{Target Geometry (Uniformity):} We regularize the set of anchors $\{\tilde{\mathbf{e}}_i\}_{i=1}^N$ to be uniformly distributed on the hypersphere. Crucially, because $\mathbf{E}$ persists across iterations, this regularization enforces constraints over the entire dataset, utilizing the full $N^2$ Gram matrix rather than just the current batch.
    \item \textbf{Learned Geometry (Alignment):} We train the encoder $f_\theta$ to maintain multi-view consistency, ensuring that different stochastic augmentations of the same sample $\mathbf{x}_i$ are mapped to nearby representations.
\end{enumerate}
To align the two objectives, we train the encoder $f_\theta$ to map the stochastic augmented views of $\mathbf{x}_i$ to its corresponding stable anchor $\tilde{\mathbf{e}}_i$.

\subsection{The IConE Objective}
\label{sec:method:objective}

The total loss function is composed of three terms, each enforcing a specific geometric constraint.\newline

\noindent\textbf{View-View Consistency ($\mathcal{L}_{\text{vv}}$).}
To ensure the representations capture the invariance defined by the augmentation pipeline, we enforce consistency between different views of the same instance:
\begin{equation}
    \mathcal{L}_{\text{vv}} = \frac{1}{B} \sum_{i \in \mathcal{B}} \frac{2}{V(V-1)} \sum_{m < n} \left(1 - \langle \mathbf{z}_i^{(m)}, \mathbf{z}_i^{(n)} \rangle \right).
\end{equation}
This term pulls views together in the representation space. Without the other terms, minimizing $\mathcal{L}_{\text{vv}}$ would lead to a trivial solution where all outputs map to a single point. Note that even if the batch size is $B=1$ (a single unique instance), as long as $V \ge 2$ augmented views are generated, this term remains well-defined.\newline

\noindent\textbf{Instance Diversity ($\mathcal{L}_{\text{div}}$).}
To prevent the geometric anchors from collapsing, we apply an explicit repulsive regularizer to the embedding table. We compute the Gram matrix $\mathbf{G} = \tilde{\mathbf{E}}\tilde{\mathbf{E}}^\top$ of the entire embedding table and minimize the off-diagonal correlations using a squared hinge loss:
\begin{equation}
    \mathcal{L}_{\text{div}} = \frac{1}{N(N-1)} \sum_{i \neq j} \left[\max(0, G_{ij})\right]^2.
\end{equation}

Since usually $N > d$ precludes strict orthogonality, we relax the constraint to penalize only positive correlations. The squared penalty makes gradients proportional to the violation, strongly repelling similar pairs while leaving nearly orthogonal ones stable. Although computing the $N \times N$ matrix is $\mathcal{O}(N^2)$, using lightweight vectors (\eg, $d=384$) keeps this compute overhead negligible relative to the backbone forward pass for our target dataset sizes. Likewise, the table's $\mathcal{O}(Nd)$ memory footprint (\eg, $\sim$7.7~MB for $N=5$k) is trivial compared to the model parameters.

However, for massive-scale datasets where computing an $N \times N$ matrix becomes a computational bottleneck, IConE's modularity provides a direct solution. Our explicit instance-to-instance repulsion can seamlessly be substituted with standard statistical anti-collapse mechanisms, such as Variance-Covariance~\cite{bardes2021vicreg} or SIGReg~\cite{balestriero2025lejepa}, applied directly to the persistent parameter space. Because these methods regularize the $d$-dimensional feature space rather than the $N \times N$ instance similarities, their computational complexity scales linearly with the dataset size. Thus, IConE offers a scalable, complementary architectural paradigm rather than a mutually exclusive competitor to existing regularization strategies.\newline

\noindent\textbf{View-Instance Alignment ($\mathcal{L}_{\text{vi}}$).}
We anchor the encoder outputs to their corresponding auxiliary instance embeddings by minimizing their cosine distance:
\begin{equation}
    \mathcal{L}_{\text{vi}} = \frac{1}{B \cdot V} \sum_{i \in \mathcal{B}} \sum_{v=1}^{V} \left(1 - \langle \mathbf{z}_i^{(v)}, \tilde{\mathbf{e}}_i \rangle \right).
\end{equation}
Since the target $\tilde{\mathbf{e}}_i$ is decoupled from the batch, this term exerts a purely attractive force on the encoder. It effectively asks the network to predict the instance's latent coordinate, which is positioned by global geometric constraints.\newline

\noindent\textbf{Total Loss.}
The complete objective is a sum:\newline
\begin{equation}
    \mathcal{L} = \mathcal{L}_{\text{vi}} + \mathcal{L}_{\text{vv}} + \mathcal{L}_{\text{div}}.
\end{equation}
where each term plays a distinct, complementary role, which we isolate empirically in Appendix F. Notably, IConE does not require temperature scaling. Contrastive SSL use temperature to control the sharpness of the softmax distribution over negatives; IConE replaces this implicit competition with direct geometric losses and eliminates this sensitive hyperparameter. The gradient structure reveals why IConE's anti-collapse mechanism is mathematically batch-agnostic. The encoder parameters $\theta$ receive gradients only from the attractive terms ($\mathcal{L}_{\text{vi}}$ and $\mathcal{L}_{\text{vv}}$). 
\begin{equation}
    \nabla_\theta \mathcal{L} = \nabla_\theta \mathcal{L}_{\text{vi}} + \nabla_\theta \mathcal{L}_{\text{vv}}.
\end{equation}
The encoder is never explicitly pushed away from other instances in the batch; it is only pulled toward its specific anchor. This is a fundamental departure from contrastive learning, where the gradient for $\mathbf{z}_i$ depends on the summation over negatives $\sum_{j \neq i} \exp(\mathbf{z}_i \cdot \mathbf{z}_j)$.

\noindent \newline The repulsive forces are handled entirely by the update to the table $\mathbf{E}$:
\begin{equation}
    \nabla_{\mathbf{e}_i} \mathcal{L} = \underbrace{\nabla_{\mathbf{e}_i} \mathcal{L}_{\text{vi}}}_{\text{attraction}} + \underbrace{\nabla_{\mathbf{e}_i} \mathcal{L}_{\text{div}}}_{\text{repulsion}}.
\end{equation}
Notably, $\mathcal{L}_{\text{vi}}$ bidirectionally updates the encoder and table $\mathbf{E}$: views are mapped toward their current anchor, while the anchor shifts toward the mean of its views. Since $\mathbf{E}$ persists across iterations, it stores the dataset's global repulsive pressure. Even at $B=1$, $\mathcal{L}_{\text{div}}$ repels $\mathbf{e}_i$ from all other embeddings $\mathbf{e}_{j \neq i}$, guaranteeing a non-collapsed target geometry without relying on in-batch negatives.

\begin{algorithm}[t]
\caption{IConE Training}
\label{alg:icone}
\begin{algorithmic}[1]
\REQUIRE Dataset $\mathcal{D}$, encoder $f_\theta$, embedding table $\mathbf{E}$, augmentation family $\mathcal{T}$
\STATE Initialize $\mathbf{E} \sim \mathcal{N}(0, 0.02)$
\FOR{each iteration}
    \STATE Sample batch $\mathcal{B}$ of $B$ instances
    \FOR{$i \in \mathcal{B}$}
        \STATE Generate $V$ views: $\mathbf{x}_i^{(v)} \sim \mathcal{T}(\mathbf{x}_i)$ for $v = 1, \ldots, V$
        \STATE $\mathbf{z}_i^{(v)} \gets \text{Normalize}(f_\theta(\mathbf{x}_i^{(v)}))$ for $v = 1, \ldots, V$
    \ENDFOR
    \STATE $\tilde{\mathbf{E}} \gets \text{RowNormalize}(\mathbf{E})$
    \STATE Compute $\mathcal{L}_{\text{vi}}$ using $\{\mathbf{z}_i^{(v)}\}$ and $\{\tilde{\mathbf{e}}_i\}_{i \in \mathcal{B}}$
    \STATE Compute $\mathcal{L}_{\text{vv}}$ using $\{\mathbf{z}_i^{(v)}\}$
    \STATE Compute $\mathcal{L}_{\text{div}}$ over full $\tilde{\mathbf{E}}$ \COMMENT{$O(N^2)$}
    \STATE $\mathcal{L} \gets \mathcal{L}_{\text{vi}} + \mathcal{L}_{\text{vv}} + \mathcal{L}_{\text{div}}$
    \STATE Update $\theta$ with learning rate $\eta$
    \STATE Update $\mathbf{E}$ with learning rate $\eta$
\ENDFOR
\end{algorithmic}
\end{algorithm}

\section{Experiments}
\label{sec:experiments}

We evaluate IConE across diverse imaging modalities, assessing its stability in small-batch and small-dataset regimes, its efficacy on high-dimensional 3D volumes, its robustness under class imbalance, and its generalization to supervised learning. We further analyze the geometric properties of the learned representations to validate our theoretical motivations. Unless otherwise noted, all results presented in this section are aggregated over all datasets within each benchmark (7 datasets for 2D and 6 datasets for 3D); per-dataset results and additional plots are provided in Appendix B.

\subsection{Experimental Setup}
\label{sec:experiments:setup}

\noindent\textbf{Training and Evaluation Protocol.}
All SSL methods are pretrained using identical encoder architectures (ViT-Small\cite{dosovitskiy2020image} for 2D, ResNet-18\cite{he2016deep} for 3D), augmentation pipelines, optimizers, learning-rate schedules, and training epochs. We evaluate learned representations via linear probing on frozen features, reporting top-1 balanced accuracy. Results are averaged over 5 random seeds, and all methods share the same dataset splits and subsampled subsets for fair comparison. Full training details and hyperparameters are provided in Appendix A.

\noindent \textbf{Datasets.}
We utilize the MedMNIST+(v2) benchmark collection~\cite{medmnistv2}, which provides standardized medical imaging datasets spanning diverse clinical domains, imaging modalities, and classification tasks. Unlike synthetic benchmarks, these datasets are derived from real clinical sources and capture the heterogeneity encountered in practice including varying image quality, and domain-specific visual patterns. This diversity makes them well-suited for evaluating method generalization across biomedical imaging applications. Each subset is sampled once and reused across all methods.

\begin{itemize}[leftmargin=*,noitemsep]
    \item \textbf{2D Benchmarks:} Blood (blood cell microscopy), Breast (ultrasound), OrganA/OrganC/OrganS (abdominal CT, different views), Path (colon pathology), and Pneumonia (chest X-ray). Image resolution of all datasets is $224 \times 224$. To simulate data-scarce clinical scenarios, we create stratified subsets with $N \in \{500, 1\text{k}, 2\text{k}, 5\text{k}\}$ total training samples.
    \item \textbf{3D Benchmarks:} Organ3D (abdominal CT), Nodule3D (chest CT), Adrenal3D (abdominal CT), Fracture3D (knee MRI), Synapse3D (electron microscopy), and Vessel3D (brain MRA). All volumes are $64^3$ voxels. These datasets represent the memory-constrained regime where small batches are often unavoidable.
\end{itemize}

\noindent \textbf{Baselines.}
We compare against contrastive~\cite{chen2020simple,he2020momentum}, clustering-based~\cite{caron2020unsupervised}, non-contrastive~\cite{grill2020bootstrap,caron2021emerging,bardes2021vicreg}, and parametric instance discrimination~\cite{balestriero2023unsupervised} baselines, using default hyperparameters for our encoders. Because standard methods require in-batch negatives or variance statistics, they inherently fail at $B=1$; thus, we report their results only where their losses are well-defined. To ensure fair comparison, we train IConE with $V=2$. Appendix C details further experiments accommodating alternative regularization methods (VCReg, SIGReg) used for IConE's diversity loss, alongside ablations on the number of views and the Gaussian initialization of instance embeddings.

\subsection{Batch-Size Stability and Performance}
\label{sec:experiments:stability}

\begin{figure}[!t]
    \centering
    \includegraphics[width=\linewidth]{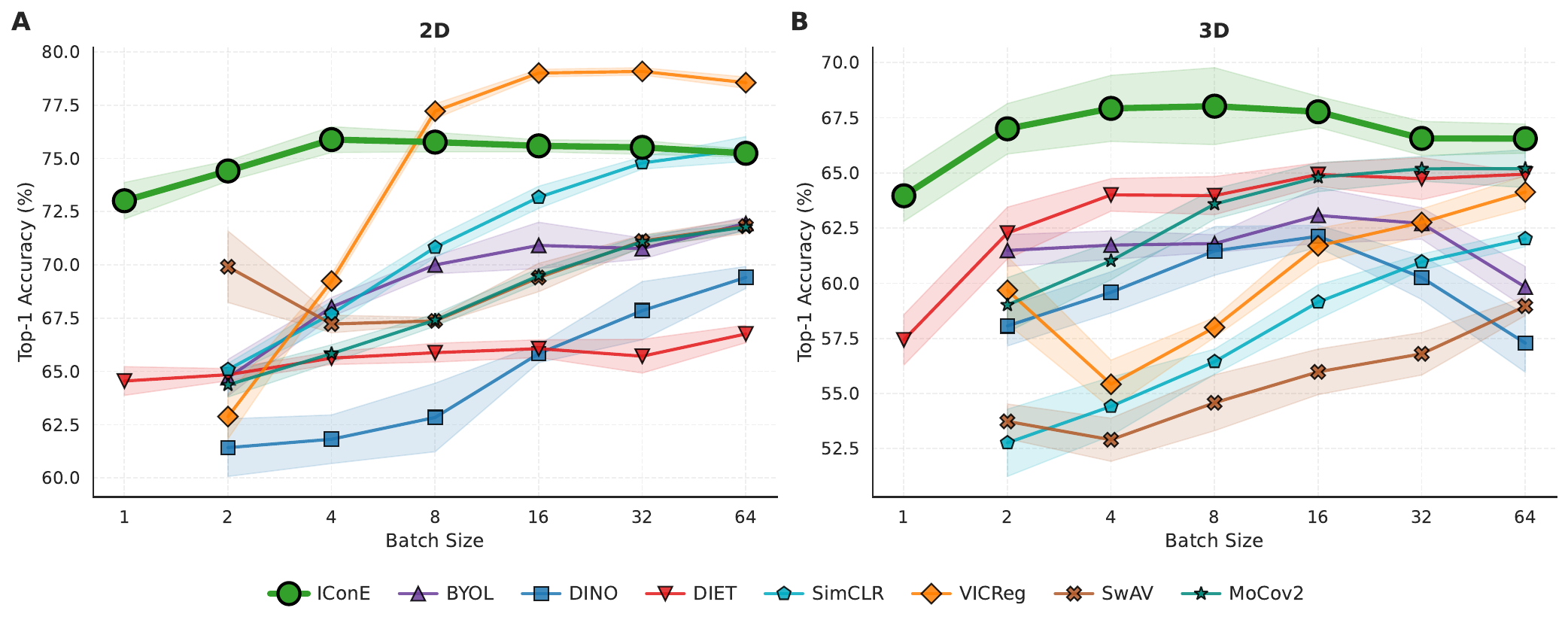}
    \caption{\textbf{Batch-size stability.} Mean and standard deviation linear-probe top-1 balanced accuracy aggregated over \textbf{(A)} 2D datasets and \textbf{(B)} 3D datasets.}
    \label{fig:main_results}
\end{figure}

We evaluate performance across batch sizes $B \in \{1, 2, 4, 8, 16, 32, 64\}$ to characterize each method's dependence on batch statistics.

\noindent\textbf{2D Results.}
\cref{fig:main_results}A presents aggregated top-1 balanced accuracy across 2D datasets. IConE exhibits stability, maintaining a flat curve. Conversely, other methods degrade characteristically as batch size decreases. While VICReg achieves the best accuracy at larger batches, it drops sharply in smaller regimes. Overall, IConE outperforms all other baselines across the spectrum, with SimCLR catching up at $B=64$, and maintains strictly higher absolute performance than DIET, which shares a parametric approach but yields lower accuracy.

\noindent\textbf{3D Results.}
\cref{fig:main_results}B demonstrates that IConE's advantages are even more pronounced in 3D, where memory constraints render small batches unavoidable. IConE consistently achieves the highest accuracy across all batch sizes. Interestingly, some baselines exhibit non-monotonic behavior, with performance dropping at the largest batch sizes, possibly showing optimization difficulties because of the use of large batch sizes in small datasets~\cite{keskar2016large}. To complement the linear probe, we evaluate the representations with a non-parametric k-NN classifier in \cref{tab:knn3d_mean_balanced_accuracy}.

\begin{table}[!t]
\centering
\caption{\textbf{Self-supervised k-NN performance.} Mean k-NN balanced accuracy (\%) across 6 3D datasets. Best results in \textbf{bold}.}
\label{tab:knn3d_mean_balanced_accuracy}
\scriptsize
\setlength{\tabcolsep}{4pt}
\begin{tabular}{@{}l ccc ccc ccc@{}}
\toprule
& \multicolumn{3}{c}{$B{=}4$} & \multicolumn{3}{c}{$B{=}16$} & \multicolumn{3}{c}{$B{=}64$} \\
\cmidrule(lr){2-4} \cmidrule(lr){5-7} \cmidrule(lr){8-10}
\textbf{Method} 
& kNN-1 & kNN-5 & kNN-20 & kNN-1 & kNN-5 & kNN-20 & kNN-1 & kNN-5 & kNN-20 \\
\midrule
BYOL    & 54.6 & 53.7 & 49.6 & 58.1 & 56.3 & 49.7 & 58.1 & 55.3 & 53.0 \\
DIET    & 55.6 & 55.6 & 54.7 & 55.3 & 55.3 & 55.3 & 57.1 & 57.0 & 54.4 \\
DINO    & 55.8 & 54.6 & 50.5 & 57.7 & 54.9 & 51.9 & 57.4 & 56.6 & 54.6 \\
MoCo-v2 & 50.8 & 51.1 & 49.7 & 56.7 & 56.2 & 55.4 & 57.2 & 56.3 & 53.3 \\
SimCLR  & 49.9 & 48.4 & 47.7 & 52.0 & 51.6 & 51.8 & 53.1 & 54.5 & 53.9 \\
SwAV    & 48.0 & 46.5 & 44.2 & 50.8 & 47.8 & 46.6 & 50.6 & 48.9 & 45.8 \\
VICReg  & 50.8 & 47.3 & 45.2 & 55.9 & 56.7 & 55.7 & 59.1 & 57.8 & 57.0 \\
\midrule
IConE (Ours) 
& \textbf{63.6} & \textbf{62.2} & \textbf{60.6} 
& \textbf{63.3} & \textbf{62.8} & \textbf{62.4} 
& \textbf{62.5} & \textbf{60.6} & \textbf{58.8} \\
\bottomrule
\end{tabular}
\end{table}

\begin{figure}[!t]
    \centering
    \includegraphics[width=\linewidth]{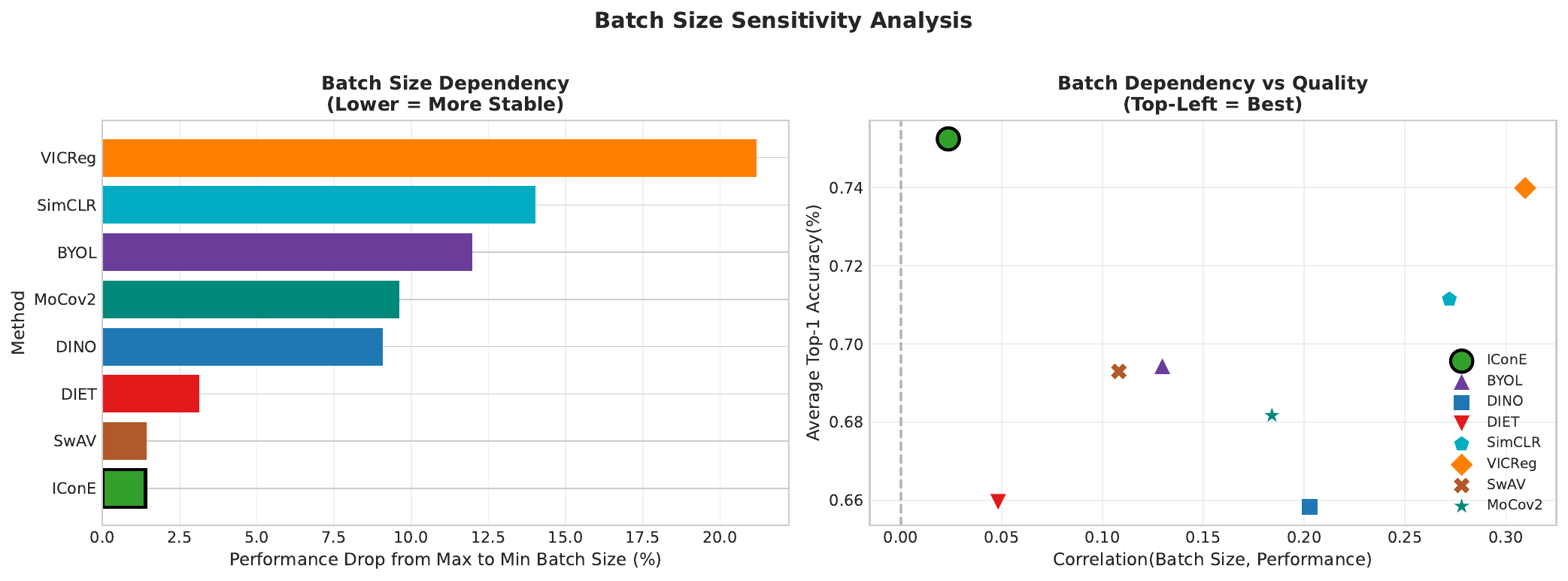}
    \caption{\textbf{Batch sensitivity analysis.} \textbf{Left:} Performance drop from largest to smallest batch size (lower is better). \textbf{Right:} Pearson Correlation between batch size and performance versus absolute top-1 balanced accuracy.}
    \label{fig:batch_sensitivity}
\end{figure}

\noindent\textbf{Batch Size Sensitivity Analysis.}
~\cref{fig:batch_sensitivity} quantifies this dependence via absolute performance drop and correlation between batch size and accuracy. The left panel shows IConE maintaining high performance with minimal degradation, avoiding the collapse of baselines. The right panel plots the trade-off between batch-size-agnosticism (low correlation, left) and representation quality (high accuracy, top). By uniquely occupying the optimal top-left corner with high accuracy and near-zero correlation, IConE achieves true batch-size-agnostic learning without sacrificing representational power. To confirm these trends extend beyond the biomedical domain, we additionally evaluate in the same small-data, small-batch regime on a natural-image subset of 20 ImageNet~\cite{deng2009imagenet} classes (250 images each); IConE exhibits the same batch-size stability there (Fig. 14, Appendix E).

\subsection{Analysis of Learned Representations}
\label{sec:experiments:repr}

We complement the evaluation with geometric analysis of the learned representations. A single linear probe evaluates only one specific task, whereas robust features must support diverse applications. A collapsed representation might therefore satisfy a specific probe yet carry limited information.

\begin{figure*}[!t]
    \centering
    \includegraphics[width=\textwidth]{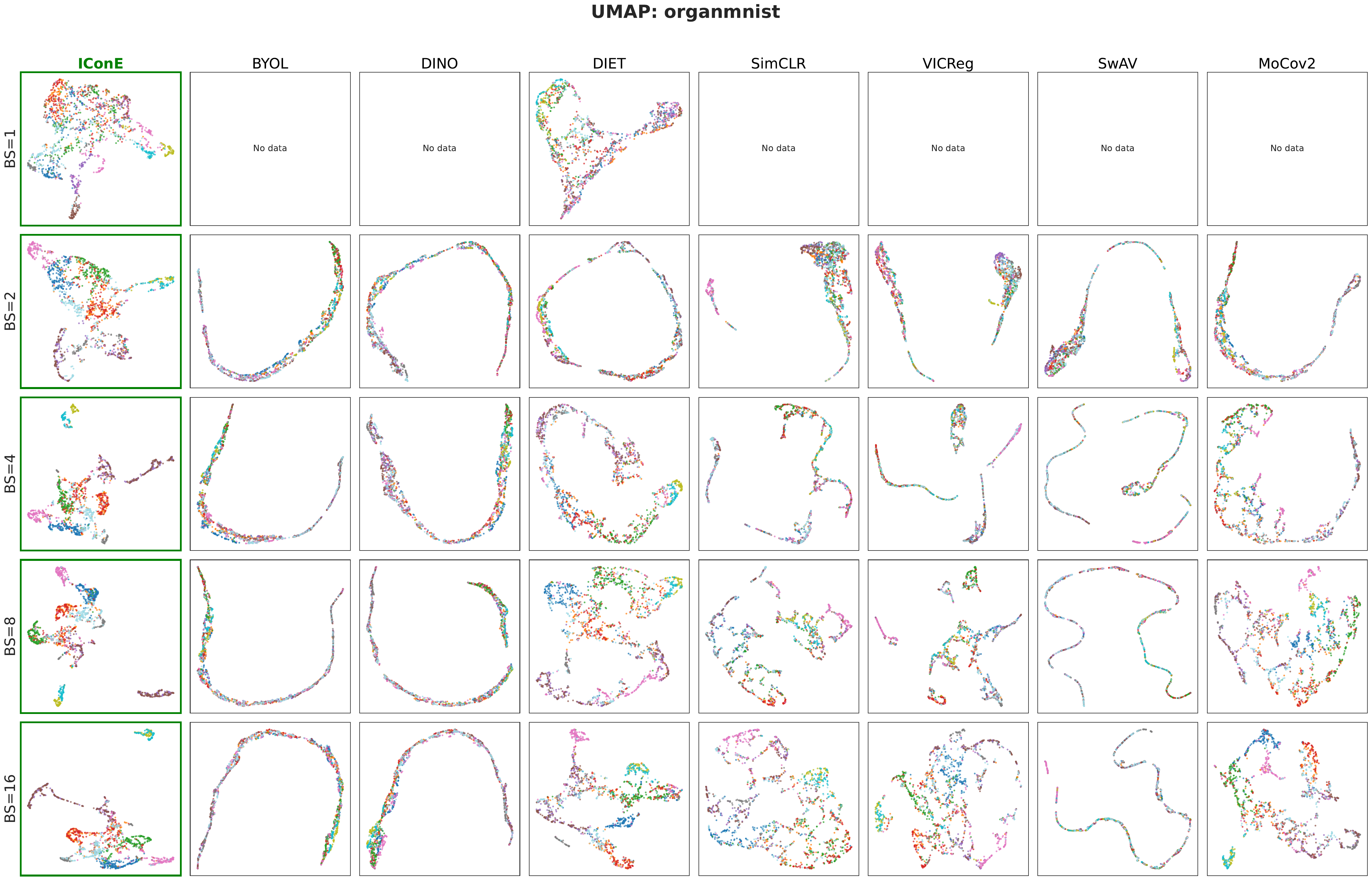}
    \caption{\textbf{UMAP visualization of learned representations.} Embeddings for OrganMNIST3D across batch sizes (rows) and methods (columns). IConE (leftmost column, green border) maintains well-separated class clusters at all batch sizes, while batch-dependent methods show progressive collapse as batch size decreases.}
    \label{fig:umap_3d}
\end{figure*}

\noindent\textbf{Qualitative Analysis.}
One intuitive way to assess representation quality is through direct visualization.~\cref{fig:umap_3d} displays UMAP projections of learned representations for OrganMNIST3D across batch sizes and methods. At larger batch sizes, some methods achieve reasonable class separation. However, as batch size decreases, baselines exhibit various forms of representational collapse, with embeddings losing their class structure. In contrast, IConE preserves well-separated, compact clusters across all batch sizes.

\begin{figure}[!t]
    \centering
    \includegraphics[width=\linewidth]{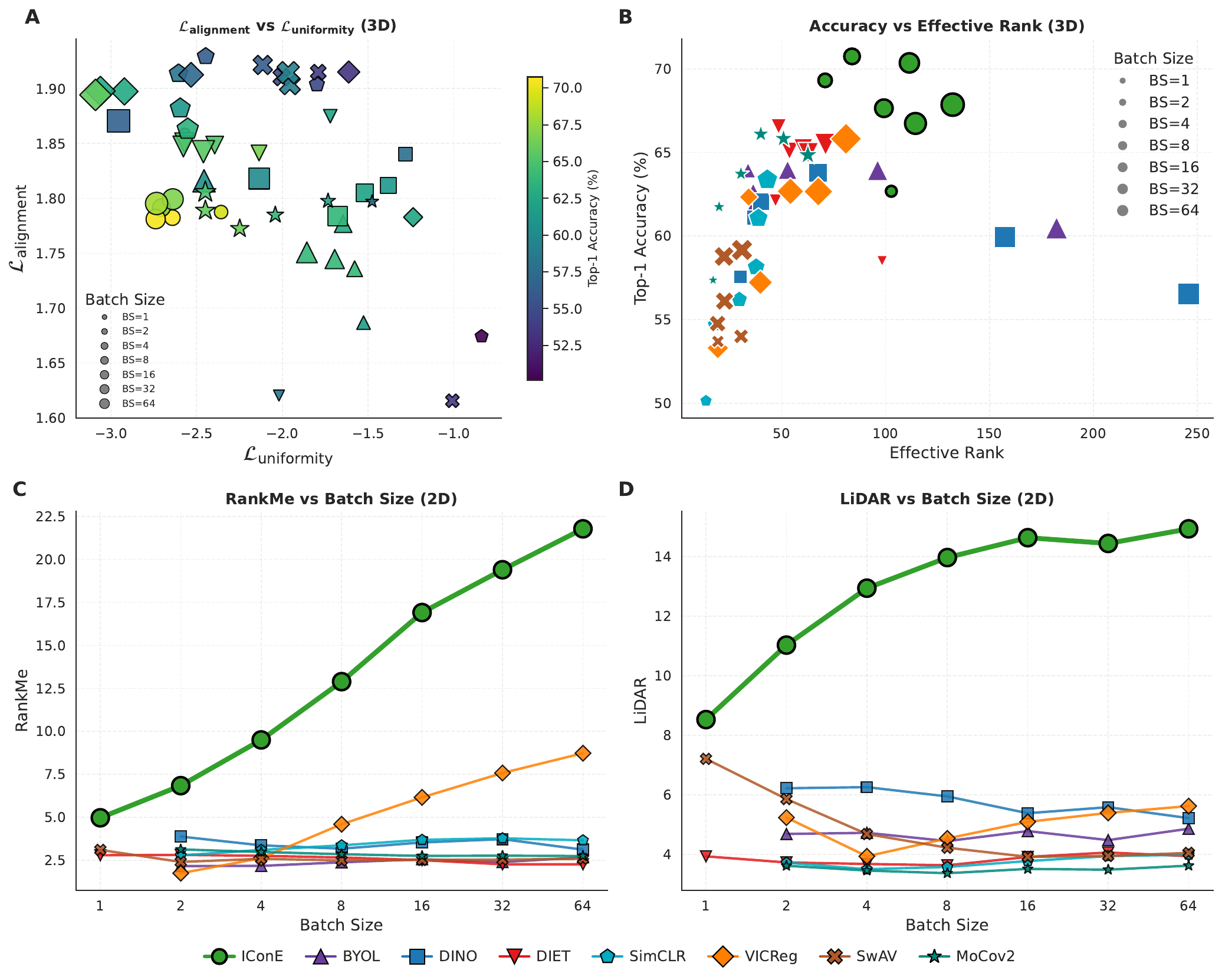}
    \caption{\textbf{Representation geometry analysis.} \textbf{(A)} Alignment vs. uniformity loss for 3D datasets, colored by downstream accuracy. \textbf{(B)} Accuracy vs. effective rank (3D). \textbf{(C)} RankMe vs. batch size (2D). \textbf{(D)} LiDAR vs. batch size (2D). IConE achieves superior geometry metrics across all measures.}
    \label{fig:repr_metrics}
\end{figure}

\noindent\textbf{Quantitative Analysis.}
We complement the qualitative analysis with established representation quality metrics. ~\cref{fig:repr_metrics} presents four complementary views.

\noindent\textbf{(A) Alignment vs. Uniformity.} Following ~\cite{wang2020understanding}, we plot alignment loss (an invariance measure) against uniformity loss (an entropy measure). Good representations should minimize both. IConE points cluster in the favorable low-alignment, low-uniformity region and are colored yellow, indicating high accuracy.

\noindent\textbf{(B) Effective Rank vs. Accuracy.} Effective rank~\cite{roy2007effective} measures how many embedding dimensions are actively utilized. We observe a strong correlation with downstream performance: the optimal top-right region represents high utilization and high accuracy. IConE consistently occupies this region across all batch sizes.

\noindent\textbf{(C--D) Collapse Indicators.} While evaluating quality without a priori information about downstream tasks is challenging, task-agnostic metrics like RankMe~\cite{garrido2023rankme} and LiDAR~\cite{thilak2023lidar} quantify dimensional collapse via the singular value spectrum. Higher values indicate greater dimensional utilization; on these metrics, IConE scores dramatically higher than all baselines across the entire batch-size range, confirming its robustness against collapse.
\subsection{Robustness to Class Imbalance}
\label{sec:experiments:imbalance}

Class imbalance poses a critical challenge for batch-dependent methods: minority classes are frequently absent from individual small batches, depriving the model of the continuous regularization signal needed to learn distinct, discriminative features for them. To test this regime, we construct class-imbalanced variants of the 2D and 3D datasets by designating half of the classes as minority classes and subsampling them by a factor $k$ (the imbalance ratio). We evaluate robustness with increasing imbalance ratios $k$, using $B=64$ to alleviate the collapse phenomenon.

\begin{figure}[!t]
    \centering
    \includegraphics[width=\linewidth]{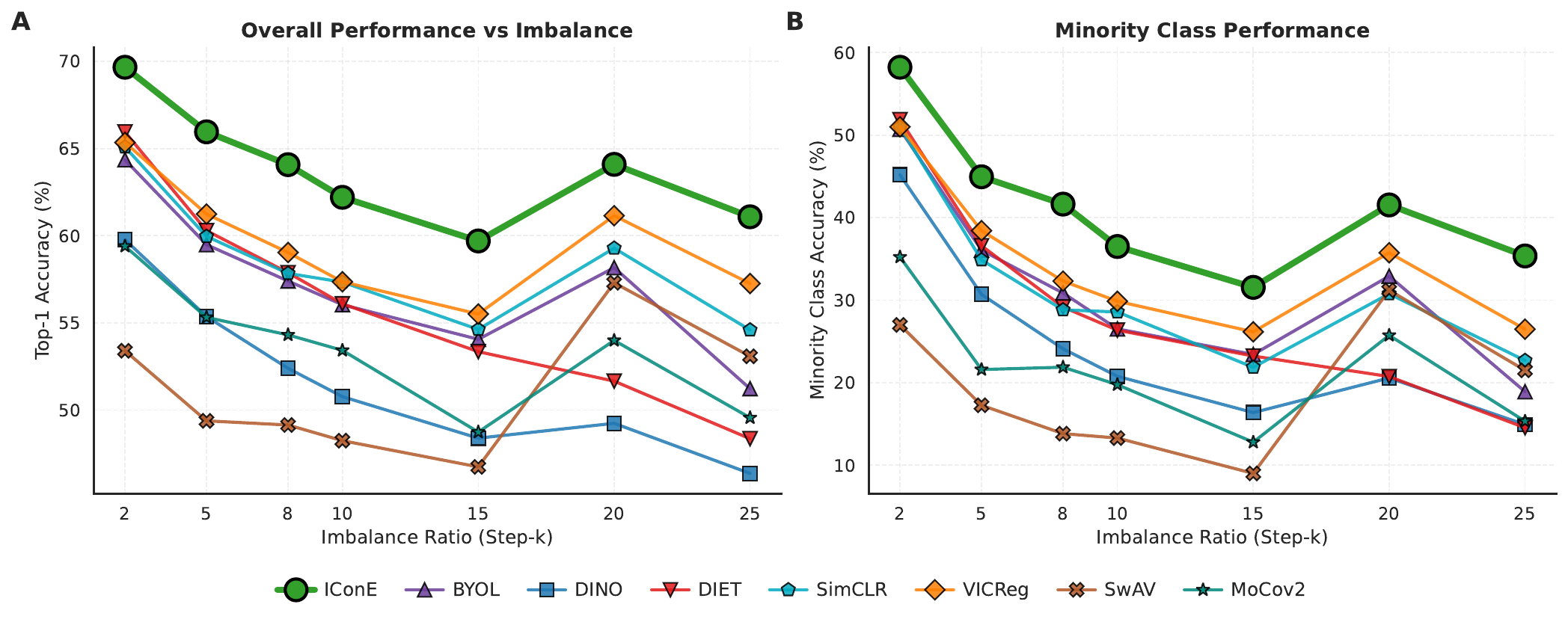}
    \caption{\textbf{Robustness to class imbalance.} \textbf{(A)} Overall top-1 balanced accuracy vs. imbalance ratio. \textbf{(B)} Minority class accuracy vs. imbalance ratio. IConE maintains superior performance on both metrics, with particularly strong advantages on minority classes at high imbalance ratios. Results are averaged over all 2D and 3D datasets.}
    \label{fig:imbalance}
\end{figure}

\noindent\textbf{Results.}
\cref{fig:imbalance}A demonstrates that IConE maintains the highest performance across all imbalance ratios. This advantage is particularly pronounced for minority classes (\cref{fig:imbalance}B), where IConE dominates baselines. This robustness stems from the persistent embedding table acting as an implicit memory: even when minority classes are absent from a batch, their corresponding anchors remain active in the diversity loss, preventing the model from forgetting underrepresented concepts.

\subsection{Extension to Supervised Learning}
\label{sec:experiments:supervised}

We investigate whether the IConE mechanism transfers to supervised learning. We compare against standard Cross-Entropy (CE) loss and Supervised Contrastive Learning (SupCon~\cite{khosla2020supervised}), and introduce two supervised variants of our method:

\begin{itemize}[leftmargin=*,noitemsep]
    \item \textbf{IConE-Class:} Replace the instance-level embedding table with a class prototype table $\mathbf{E} \in \mathbb{R}^{C \times d}$. The encoder aligns views to their class prototype, while prototypes are regularized for orthogonality.
    \item \textbf{IConE-Instance:} Retain instance-level anchors but use labels: the alignment loss pulls views toward same-class anchors, while the diversity loss repels cross-class anchors.
\end{itemize}


\begin{table}[!t]
\centering
\caption{\textbf{Supervised learning performance.} Aggregated top-1 balanced accuracy (\%) across batch sizes for 2D and 3D datasets. Best in \textbf{bold}, second best \underline{underlined}.}
\label{tab:supervised}
\scriptsize
\setlength{\tabcolsep}{2.5pt}
\begin{tabular}{@{}l cccccc cccccc@{}}
\toprule
& \multicolumn{6}{c}{\textbf{2D}} & \multicolumn{6}{c}{\textbf{3D}} \\
\cmidrule(lr){2-7} \cmidrule(lr){8-13}
\textbf{Method} 
& $B{=}1$ & $B{=}4$ & $B{=}8$ & $B{=}16$ & $B{=}32$ & $B{=}64$ 
& $B{=}1$ & $B{=}4$ & $B{=}8$ & $B{=}16$ & $B{=}32$ & $B{=}64$ \\
\midrule
Cross-Entropy   
& 76.20 & 78.17 & \underline{79.72} & \underline{79.87} & \underline{80.67} & \underline{80.32}
& 66.61 & 72.70 & 71.81 & 69.65 & 73.03 & 72.85 \\
SupCon          
& 72.20 & 74.08 & 76.32 & 79.70 & 79.93 & 79.73
& 59.49 & 66.50 & \underline{72.97} & 72.59 & 67.99 & 71.55 \\
\midrule
IConE-Instance  
& \underline{76.81} & \underline{78.43} & 78.56 & 79.53 & 79.58 & 79.63
& \underline{68.77} & \textbf{75.16} & \textbf{73.14} & \underline{76.56} & \textbf{75.10} & \underline{75.31} \\
IConE-Class     
& \textbf{78.71} & \textbf{82.42} & \textbf{81.97} & \textbf{82.16} & \textbf{82.28} & \textbf{82.31}
& \textbf{69.59} & \underline{74.25} & 72.13 & \textbf{77.14} & \underline{74.34} & \textbf{76.21} \\
\bottomrule
\end{tabular}
\end{table}

\noindent\textbf{Results.}
\cref{tab:supervised} shows both IConE variants outperform SupCon across all batch sizes, as SupCon inherits the same batch dependency as its unsupervised counterpart (SimCLR) due to reliance on in-batch positives. More notably, IConE-Class surpasses even standard CE. We hypothesize this stems from CE's reliance on prolonged training to converge toward a maximally separated class geometry~\cite{papyan2020prevalence}, a phenomenon known as Neural Collapse. IConE-Class circumvents this by directly imposing inter-class separation via $\mathcal{L}_{\text{div}}$ and augmentation invariance via $\mathcal{L}_{\text{vv}}$, neither of which has an analogue in standard CE training.

\subsection{Modularity and Scalability of $\mathcal{L}_{\text{div}}$}
\label{sec:modularity}

The diversity term $\mathcal{L}_{\text{div}}$ is modular: the default $\mathcal{O}(N^2)$ orthogonality penalty can be replaced by any anti-collapse mechanism on the embedding table. \cref{tab:main_ablation_regularizer} shows that substituting VCReg~\cite{bardes2021vicreg} or SIGReg~\cite{balestriero2025lejepa} regularization leaves accuracy within $\sim$1--2 points across batch sizes, confirming the mechanism is agnostic to the form of $\mathcal{L}_{\text{div}}$, while scaling much better in terms of GPU memory (Appendix C).

\begin{table}[!t]
\centering
\caption{\textbf{Regularizer ablation.} Top-1 accuracy (\%) for IConE with default orthogonality, VCReg, SIGReg regularization on 2D and 3D datasets across batch sizes.}
\label{tab:main_ablation_regularizer}
\scriptsize
\setlength{\tabcolsep}{2.5pt}
\begin{tabular}{@{}l cccccc cccccc@{}}
\toprule
& \multicolumn{6}{c}{\textbf{2D}} & \multicolumn{6}{c}{\textbf{3D}} \\
\cmidrule(lr){2-7} \cmidrule(lr){8-13}
\textbf{Regularizer}
& $B{=}1$ & $B{=}2$ & $B{=}4$ & $B{=}8$ & $B{=}16$ & $B{=}32$
& $B{=}1$ & $B{=}2$ & $B{=}4$ & $B{=}8$ & $B{=}16$ & $B{=}32$ \\
\midrule
IConE (default) & 59.6 & 59.5 & 60.6 & 60.2 & 62.0 & 60.1 & 62.1 & 67.9 & 69.7 & 66.7 & 68.8 & 65.4 \\
IConE-VCReg     & 59.3 & 60.2 & 60.8 & 60.9 & 60.4 & 61.2 & 63.0 & 66.1 & 68.8 & 69.2 & 69.1 & 67.9 \\
IConE-SIGReg    & 57.3 & 58.4 & 58.3 & 58.5 & 58.5 & 58.4 & 62.9 & 67.5 & 68.4 & 68.9 & 67.5 & 68.0 \\
\bottomrule
\end{tabular}
\end{table}

\section{Conclusion}
\label{sec:conclusion}

While self-supervised learning is highly effective, standard methods rely heavily on batch statistics, causing performance degradation at the small batch sizes necessitated by high-dimensional data. To resolve this, we introduced \textbf{IConE}, a batch-size-robust approach that replaces dynamic batch interactions with a persistent, learnable embedding table. By aligning augmented views to these stable anchors and applying an explicit orthogonality regularizer, IConE prevents representation collapse independently of batch composition, maintaining consistent performance from batch size 1 to 64. Empirically, IConE achieves state-of-the-art small-batch performance across diverse 2D and 3D tasks, with geometric analysis confirming it successfully prevents dimensional collapse. Furthermore, the framework demonstrates superior robustness to class imbalance, and its core mechanism generalizes effectively to supervised representation learning.\newline

\noindent\textbf{Limitations and Future Work.} While standard methods may outperform IConE when massive batches are computationally feasible, high-dimensional scientific data fundamentally restricts batch size, making our small-batch focus practically necessary. Moreover, eliminating the need for large batches frees significant memory, enabling the training of much larger encoder architectures. Regarding dataset scalability, while our full-table orthogonality loss scales as $\mathcal{O}(N^2)$, this overhead is negligible in our target domains. For massive-scale future applications, IConE's modularity accommodates $\mathcal{O}(N \cdot d^2)$ statistical regularizers without compromising its core decoupling principle. Ultimately, by removing large-batch dependencies, IConE provides a robust, scalable path forward for high-dimensional self-supervised learning.

\section*{Acknowledgements}
RT-SuperES has received funding from the European Innovation Council (EIC)
under grant agreement No.~101099654.

%
%
\bibliographystyle{splncs04}
\bibliography{main}

\clearpage
\appendix

\section{Training Details}
\label{sec:appendix:training}

We provide complete implementation details to ensure reproducibility. All experiments were conducted using PyTorch.

\subsection{Architecture Specifications}

\noindent\textbf{2D Experiments.}
We use a Vision Transformer Small (ViT-S) backbone with the following configuration: patch size $16 \times 16$, embedding dimension $d=384$, depth 12, 6 attention heads, and MLP ratio 4. Input images are resized to $224 \times 224$.\newline 

\noindent\textbf{3D Experiments.}
We use a 3D ResNet-18 backbone adapted for volumetric inputs. The architecture follows the standard ResNet-18 structure with all 2D convolutions replaced by 3D counterparts. The final fully-connected layer projects to embedding dimension $d=512$. Input volumes are $64 \times 64 \times 64$ voxels with a single channel.

\subsection{Augmentation Pipelines}
\label{sec:appendix:augmentations}

\noindent\textbf{2D Augmentations.}
We apply the standard SSL augmentation pipeline used by SimCLR and related methods:
\begin{itemize}[leftmargin=*,noitemsep]
    \item RandomResizedCrop to $224 \times 224$ with scale range $[0.2, 1.0]$
    \item RandomHorizontalFlip with probability $p=0.5$
    \item ColorJitter (brightness=0.4, contrast=0.4, saturation=0.4, hue=0.1) with $p=0.8$
    \item RandomGrayscale with $p=0.2$
    \item GaussianBlur with kernel size 23 and $\sigma \in [0.1, 2.0]$
    \item Normalization with ImageNet statistics (mean=$(0.485, 0.456, 0.406)$,
    
    std=$(0.229, 0.224, 0.225)$)
\end{itemize}

\noindent\textbf{3D Augmentations.}
We design a 3D-specific augmentation pipeline suitable for volumetric medical data:
\begin{itemize}[leftmargin=*,noitemsep]
    \item Random 3D resized crop with scale range $[0.5, 1.0]$, resized to $64^3$ via trilinear interpolation
    \item Random flips along each spatial axis (D, H, W) independently with $p=0.5$
    \item Random 90-degree rotations around a randomly selected axis pair with $p=0.5$
    \item Intensity shift: additive offset sampled uniformly from $[-0.1, 0.1]$
    \item Contrast adjustment: multiplicative scaling sampled uniformly from $[0.8, 1.2]$
    \item Gaussian noise with $\sigma=0.1$ applied with $p=0.3$
    \item Gaussian blur (approximated via 3D average pooling with kernel size 3) with $p=0.3$
\end{itemize}

\subsection{Optimization Hyperparameters}

\cref{tab:hyperparams} summarizes the optimization settings. Notably, we run IConE experiments with no loss weighting hyperparameters.

\begin{table}[h]
\centering
\caption{\textbf{Optimization hyperparameters.} IConE uses identical settings across 2D and 3D, with the only difference being embedding dimension (determined by the backbone).}
\label{tab:hyperparams}
\begin{tabular}{@{}lcc@{}}
\toprule
\textbf{Hyperparameter} & \textbf{2D} & \textbf{3D} \\
\midrule
Optimizer & AdamW & AdamW \\
Base learning rate & $1 \times 10^{-4}$ & $1 \times 10^{-4}$ \\
Weight decay & 0.05 & 0.05 \\
LR schedule & Cosine annealing & Cosine annealing \\
Training epochs & 100 & 100 \\
Number of views $V$ & 2 & 2 \\
Embedding dimension $d$ & 384 & 512 \\
Embedding initialization & $\mathcal{N}(0, 0.02)$ & $\mathcal{N}(0, 0.02)$ \\
\bottomrule
\end{tabular}
\end{table}

\subsection{Baseline Implementation Details}

All baselines use identical encoder architectures, augmentation pipelines, optimizers, and training durations as IConE to ensure fair comparison. We use the following hyperparameters:

\begin{itemize}[leftmargin=*,noitemsep]
    \item \textbf{SimCLR}: Temperature $\tau=0.5$, 2-layer MLP projector with hidden dimension 2048 and output dimension 128. Undefined at $B=1$ (requires in-batch negatives).
    \item \textbf{MoCo v2}: Queue size 4096, momentum coefficient $m=0.99$, temperature $\tau=0.2$.
    \item \textbf{SwAV}: 300 prototypes, 3 Sinkhorn-Knopp iterations, temperature 0.1. Requires $B \geq 2$ for meaningful cluster assignments.
    \item \textbf{BYOL}: Momentum coefficient $m=0.99$, 2-layer MLP projector and predictor with hidden dimension 4096.
    \item \textbf{VICReg}: Invariance weight $\lambda=25$, variance weight $\mu=25$, covariance weight $\nu=1$.
    \item \textbf{DINO}: Teacher momentum 0.996$\to$1.0 (cosine schedule), student temperature 0.1, teacher temperature 0.04$\to$0.07.
    \item \textbf{DIET}: Label smoothing factor 0.8, softmax-based instance discrimination.
\end{itemize}

\subsection{Evaluation Protocol}

\noindent\textbf{Linear Probing.}
We evaluate learned representations by training a linear classifier on frozen features using scikit-learn's LogisticRegression with L-BFGS optimizer (max 1000 iterations), multinomial cross-entropy loss, and default L2 regularization. Features are standardized to zero mean and unit variance before training. We report top-1 balanced accuracy to account for class imbalance in the datasets.\newline

\noindent\textbf{Data Splits.}
For each dataset size configuration ($N \in \{500, 1\text{k}, 2\text{k}, 5\text{k}\}$ for 2D, full dataset for 3D), we perform stratified sampling from the original MedMNIST training data to create our training subsets. We then split each subset into train/test partitions. Crucially, the same subsets and splits are reused across all methods and random seeds to ensure fair comparison. All reported results are averaged over 5 random seeds with different model initializations.

\section{Per-Dataset Results}
\label{sec:appendix:per_dataset}

The main paper reports results aggregated across all datasets within each benchmark. Here we provide detailed per-dataset performance to demonstrate the consistency of IConE's advantages across diverse imaging modalities and classification tasks.

\subsection{2D MedMNIST Datasets}

\begin{figure*}[!htbp]
    \centering
    \includegraphics[width=\textwidth]{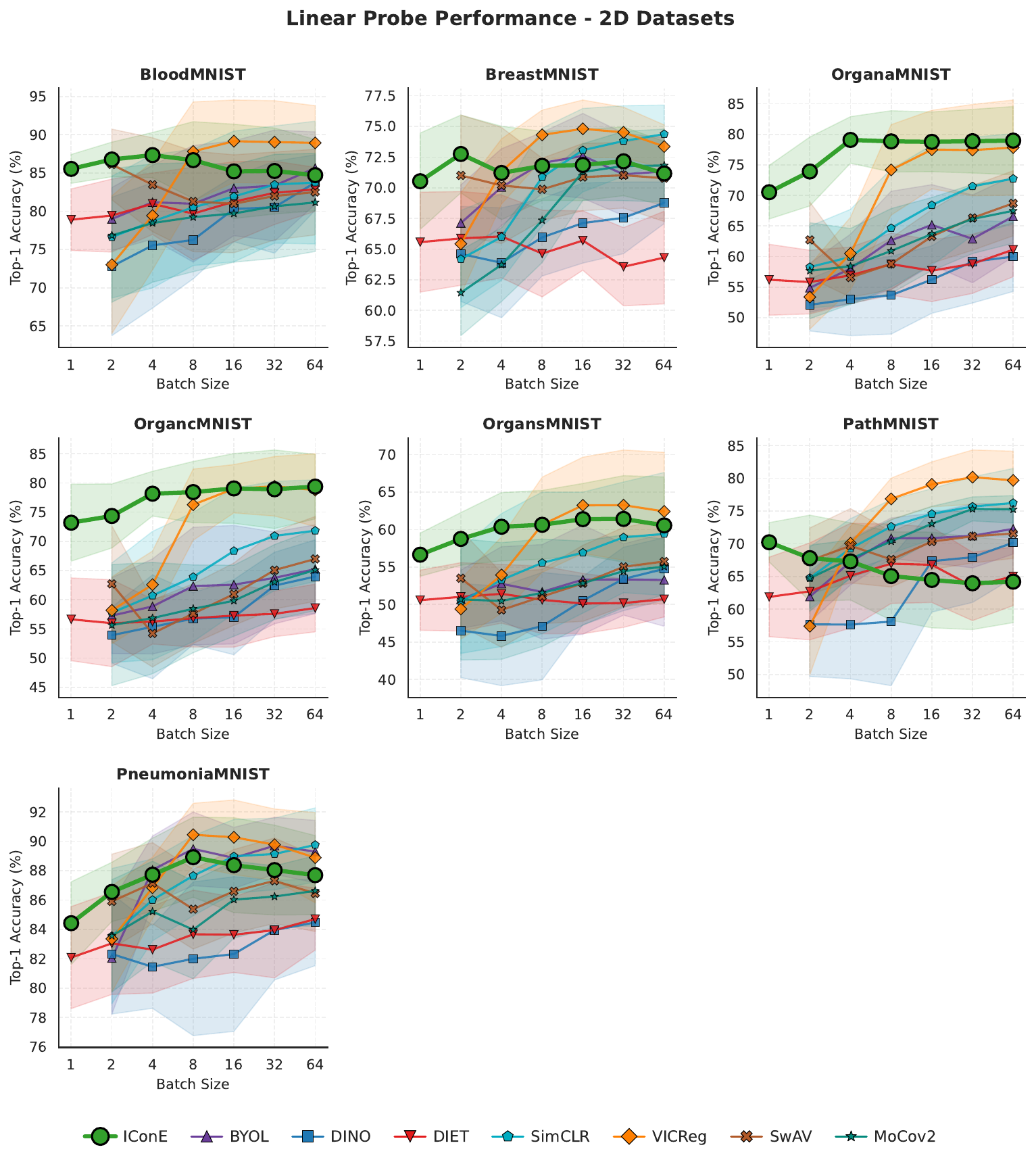}
    \caption{\textbf{Per-dataset 2D results.} Linear probe balanced accuracy vs.\ batch size for each 2D MedMNIST dataset. In most datasets, IConE maintains consistent performance across all batch sizes, while other baselines show characteristic degradation patterns at small batch sizes.}
    \label{fig:per_dataset_2d}
\end{figure*}

\cref{fig:per_dataset_2d} presents linear probe top-1 balanced accuracy across batch sizes for each 2D dataset individually. IConE maintains stable performance across all batch sizes on almost every dataset, while baselines exhibit varying degrees of degradation as batch size decreases.

\subsection{3D MedMNIST Datasets}

\begin{figure*}[!htbp]
    \centering
    \includegraphics[width=\textwidth]{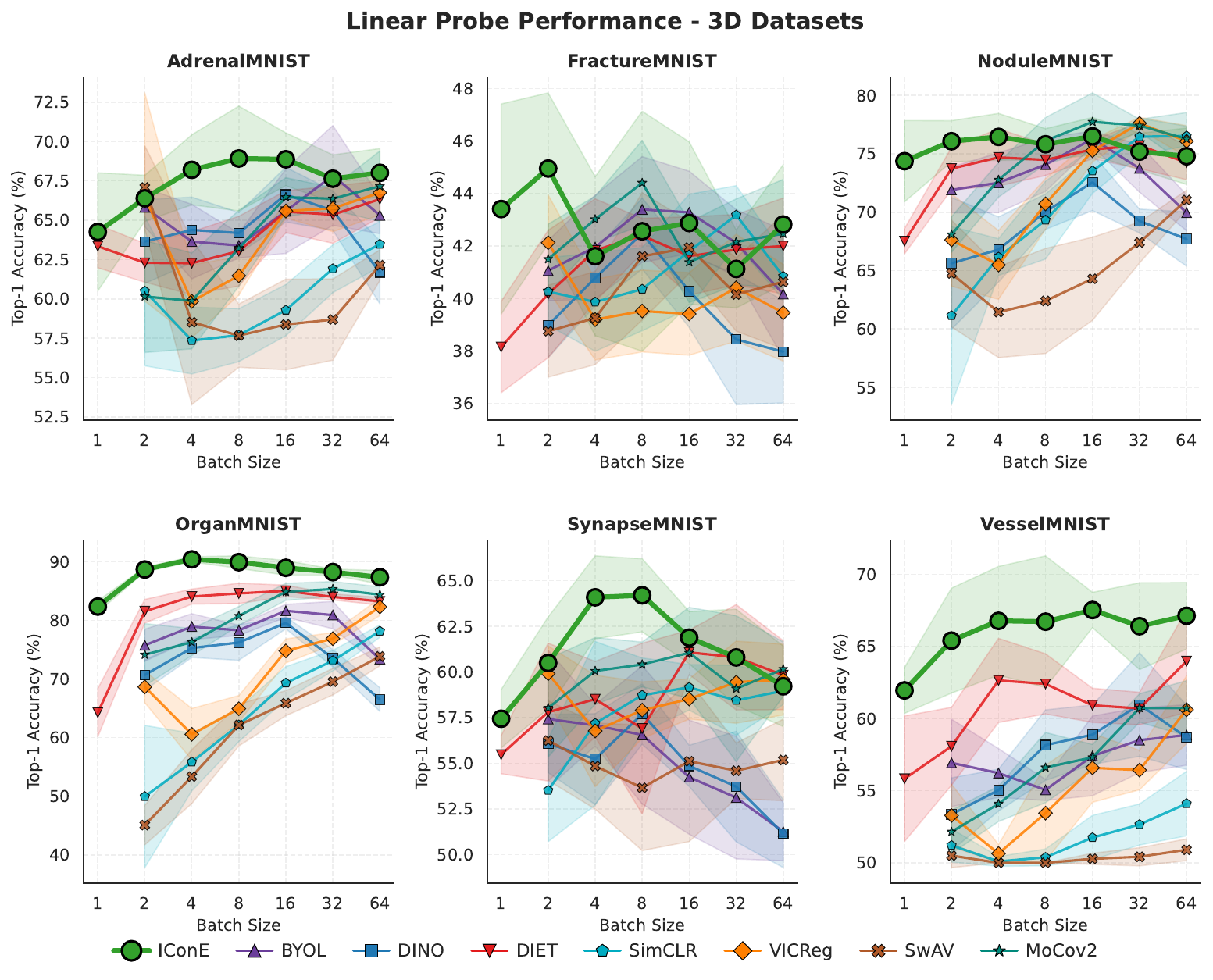}
    \caption{\textbf{Per-dataset 3D results.} Linear probe balanced accuracy vs.\ batch size for each 3D MedMNIST dataset. IConE's batch-agnostic design provides consistent advantages across all volumetric imaging tasks.}
    \label{fig:per_dataset_3d}
\end{figure*}

\cref{fig:per_dataset_3d} shows corresponding results for the 3D volumetric datasets. IConE matches or surpasses other baselines in most datasets across different batch sizes.

\subsection{Joint Batch Size and Dataset Size Analysis}

\begin{figure*}[t]
    \centering
    \includegraphics[width=\textwidth]{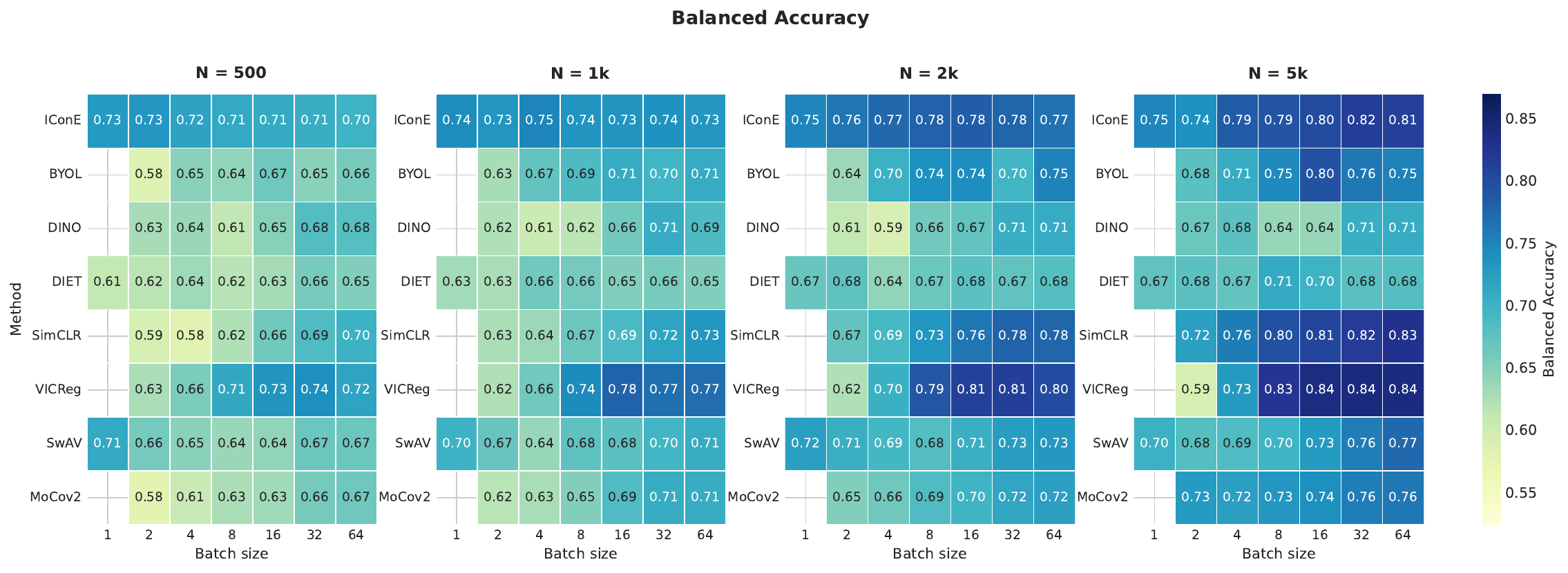}
    \caption{\textbf{Joint batch size and dataset size analysis.} Balanced accuracy heatmaps for dataset sizes $N \in \{500, 1\text{k}, 2\text{k}, 5\text{k}\}$. IConE maintains stable high performance (dark blue) across all conditions, while baselines degrade significantly at small batch sizes, particularly in data-scarce regimes.}
    \label{fig:heatmap_batch_dataset}
\end{figure*}

\cref{fig:heatmap_batch_dataset} presents a comprehensive view of performance across both batch sizes and dataset sizes simultaneously. Each heatmap shows balanced accuracy for a given dataset size $N$, with methods on the vertical axis and batch sizes on the horizontal axis. IConE (top row) maintains consistently high accuracy (dark blue) across all conditions, while other methods show pronounced degradation in the top-left region (small batch, small dataset), which represents the most challenging and practically relevant regime for scientific imaging.

\section{Ablation Studies}
\label{sec:appendix:ablations}

We conduct ablation studies to analyze IConE's sensitivity to design choices and hyperparameters.

\subsection{Alternative Regularization Strategies}
\label{sec:ablation_regularizer}

As noted in the main text, IConE's diversity penalty is modular. The default orthogonality regularizer can be replaced with alternative anti-collapse mechanisms. We evaluate two alternatives: variance-covariance (VC) regularization and Sketched Isotropic Gaussian (SIG) regularization.\newline

\noindent\textbf{Variance-Covariance Regularization.} We apply variance-covariance regularization to the instance embedding table:
\begin{equation}
    \mathcal{L}_{\text{vc}} = \lambda_{\text{var}} \sum_{j=1}^{d} \max\left(0, \gamma - \sqrt{\text{Var}(\tilde{\mathbf{e}}_{:,j}) + \epsilon}\right) + \lambda_{\text{cov}} \sum_{i \neq j} \text{Cov}(\tilde{\mathbf{e}}_{:,i}, \tilde{\mathbf{e}}_{:,j})^2
\end{equation}
where $\gamma=1$ is the target standard deviation, $\epsilon=10^{-4}$, and we use $\lambda_{\text{var}}=\lambda_{\text{cov}}=1$. The variance term prevents dimensional collapse by ensuring each embedding dimension maintains sufficient variance, while the covariance term decorrelates dimensions.\newline

\noindent\textbf{Sketched Isotropic Gaussian Regularization (SIGReg).}
Following SIGReg, we regularize the full persistent instance embedding table
$E \in \mathbb{R}^{N \times d}$ using random one-dimensional projections and a Gaussian characteristic-function discrepancy.
After per-dimension standardization of $E$ to obtain $\tilde E$, we sample $M$ random unit vectors
$\{\mathbf{a}_m\}_{m=1}^M$ and form projected samples
$z_i^{(m)} = \mathbf{a}_m^\top \tilde{\mathbf{e}}_i$.
For each projection, we compute the empirical characteristic function
\[
\hat{\phi}_m(t)=\frac{1}{N}\sum_{i=1}^{N} e^{\,it z_i^{(m)}}
\]
and compare it to the characteristic function of $\mathcal{N}(0,1)$,
$\phi_G(t)=e^{-t^2/2}$, via
\[
T_m
=
N \int_{-R}^{R}
\left|
\hat{\phi}_m(t)-\phi_G(t)
\right|^2
\phi_G(t)\,dt.
\]
The integral is approximated numerically on a finite grid, and the final loss is
\[
\mathcal{L}_{\text{sig}}=\frac{1}{M}\sum_{m=1}^{M} T_m.
\]
This encourages the persistent instance embeddings to exhibit Gaussian behavior along many random directions, promoting an approximately isotropic and non-collapsed global embedding geometry.\newline


\noindent\textbf{Results.} Main text Tab. 3 compares the three regularization strategies across batch sizes on both 2D and 3D MedMNIST datasets. All three approaches achieve comparable performance, with the default orthogonality regularizer providing a slight edge at smaller batch sizes. This confirms that IConE's effectiveness stems primarily from the architectural decoupling of alignment and uniformity objectives (regularization is applied to the full embedding table and not to mini-batches), rather than the specific choice of regularizer.


\subsection{Number of Views}

We evaluate the effect of the number of augmented views $V \in \{2, 4, 8, 12\}$ generated per instance during training. \cref{tab:ablation_views} shows results across different batch sizes. Performance is relatively stable across view counts, with $V=2$ already achieving strong results. While additional views provide marginal improvements in certain configurations, they increase memory consumption linearly. Given that our target regime is memory-constrained, we use $V=2$ as the default, which balances performance and efficiency.

\begin{table}
\centering
\caption{\textbf{Number of views ablation.} Top-1 Balanced accuracy (\%) on 2D datasets across various batch sizes. $V=2$ provides strong average performance with minimal memory overhead.}
\label{tab:ablation_views}
\begin{tabular}{@{}l cccccc c@{}}
\toprule
 & \multicolumn{6}{c}{\textbf{Batch Size ($B$)}} & \\
\cmidrule(lr){2-7}
\textbf{Views ($V$)} & $1$ & $2$ & $4$ & $8$ & $16$ & $32$ & \textbf{Average} \\
\midrule
$2$ (default) & 73.4 & 74.7 & 76.1 & 75.8 & 75.5 & 75.3 & 75.1 \\
$4$  & 72.6 & 74.8 & 76.1 & 76.1 & 76.0 & 75.4 & 75.1 \\
$8$  & 72.8 & 74.6 & 76.1 & 75.9 & 76.1 & 75.4 & 75.2 \\
$12$ & 72.1 & 74.5 & 75.9 & 75.8 & 75.2 & 75.2 & 74.8 \\
\bottomrule
\end{tabular}
\end{table}

\subsection{Embedding Initialization Scale}

The embedding table $\mathbf{E}$ is initialized from a normal distribution $\mathcal{N}(0, \sigma)$. We ablate the initialization scale $\sigma \in \{0.02, 0.05, 0.10, 0.20, 0.50\}$ to assess sensitivity. \cref{tab:ablation_init} shows that performance is robust across a wide range of initialization scales, with $\sigma=0.02$ providing slightly better results for $B=1$.

\begin{table}
\centering
\caption{\textbf{Initialization scale ablation.} Balanced accuracy (\%) on 2D datasets across various batch sizes. Performance remains robust across initialization scales spanning an order of magnitude.}
\label{tab:ablation_init}
\begin{tabular}{@{} l cccccc c @{}}
\toprule
& \multicolumn{6}{c}{\textbf{Batch Size ($B$)}} & \\
\cmidrule(lr){2-7}
\textbf{Init scale ($\sigma$)} & $1$ & $2$ & $4$ & $8$ & $16$ & $32$ & \textbf{Average} \\
\midrule
$0.02$ (default) & 74.3 & 74.2 & 75.8 & 75.5 & 75.7 & 76.0 & 75.2 \\
$0.05$           & 72.3 & 75.5 & 76.0 & 76.2 & 75.1 & 76.1 & 75.2 \\
$0.10$           & 71.3 & 75.4 & 75.6 & 75.9 & 75.7 & 75.9 & 75.0 \\
$0.20$           & 70.9 & 74.9 & 76.8 & 75.3 & 76.0 & 75.5 & 74.9 \\
$0.50$           & 72.2 & 75.6 & 76.8 & 76.3 & 75.0 & 75.0 & 75.1 \\
\bottomrule
\end{tabular}
\end{table}

\subsection{GPU Memory and Wall-Clock Time}
\label{sec:supp_memory_time}

\begin{figure}[h]
    \centering
    \includegraphics[width=\linewidth]{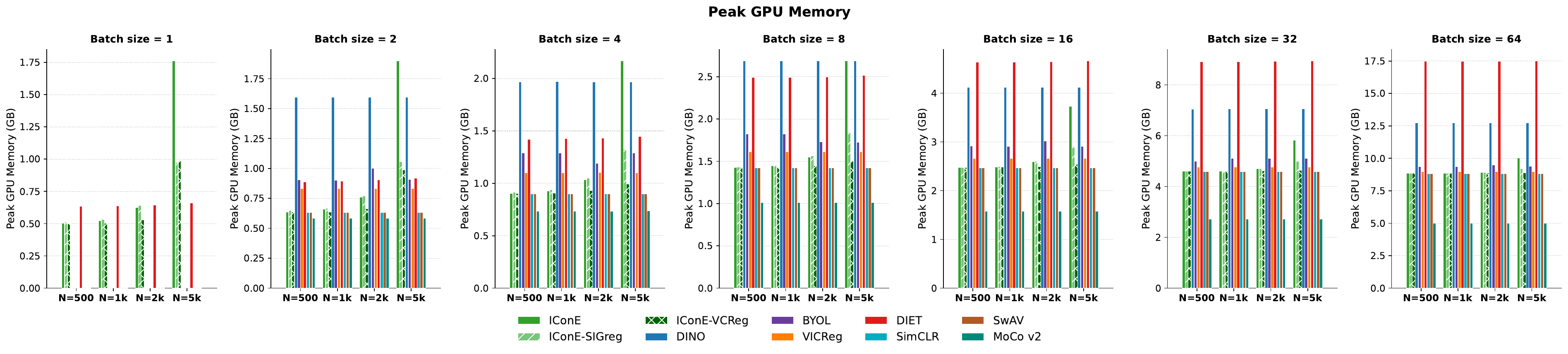}
    \caption{
        Peak GPU memory (GB) across batch sizes and dataset sizes for all evaluated SSL methods on BloodMNIST.
        Each panel corresponds to a different batch size.
        IConE-SIGreg and IConE-VCReg match the memory footprint of competitive baselines across all settings,
        whereas the default IConE variant incurs increasing overhead as $N$ grows.
    }
    \label{fig:supp_gpu_memory_sweep}
\end{figure}

\begin{figure}[h]
    \centering
    \includegraphics[width=\linewidth]{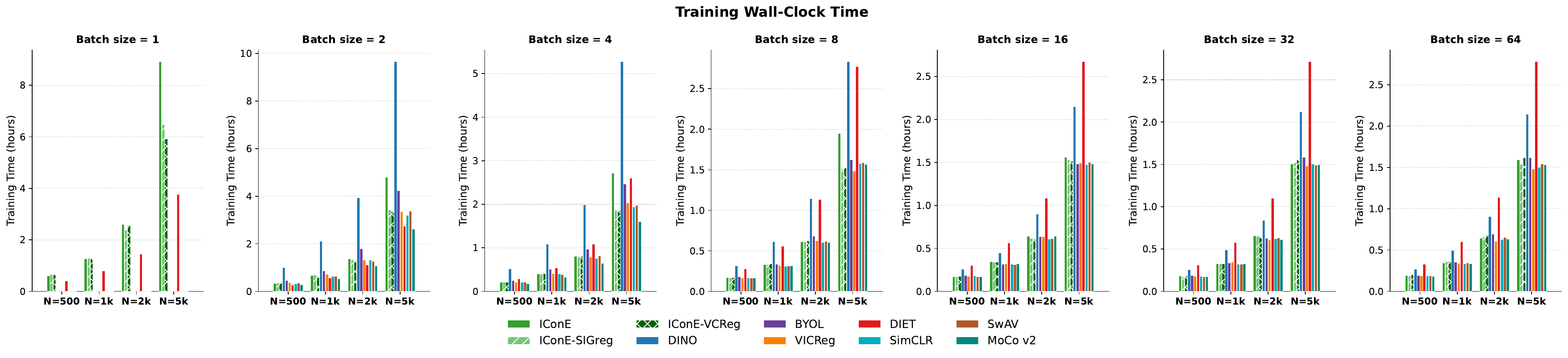}
    \caption{
        Wall-clock training time (hours) across batch sizes and dataset sizes for all evaluated SSL methods on BloodMNIST.
        The default IConE variant exhibits super-linear growth in training time with dataset size due to its $\mathcal{O}(N^2)$ diversity regularisation, while IConE-SIGreg and IConE-VCReg scale comparably to the remaining baselines.
    }
    \label{fig:supp_wallclock_sweep}
\end{figure}

To assess the computational cost of IConE and its variants relative to competing SSL methods, we measure peak GPU memory allocation and wall-clock training time across dataset sizes ($N \in \{500, 1\text{k}, 2\text{k}, 5\text{k}\}$) and batch sizes ($\{1, 2, 4, 8, 16, 32, 64\}$) on a 2D dataset.
All experiments run on a single NVIDIA A40 GPU. Results are shown in ~\cref{fig:supp_gpu_memory_sweep,fig:supp_wallclock_sweep}.\newline

\noindent\textbf{Scaling behaviour.}
Most baselines exhibit roughly constant memory and time costs as $N$ grows, since their loss functions operate exclusively in a fixed-dimensional embedding space and carry no parameters tied to dataset size.
The default IConE variant deviates from this pattern: both memory and training time increase noticeably with $N$, reflecting the cost of its instance diversity regularisation term, which computes a full $N \times N$ pairwise similarity matrix over all instance embeddings at every training step.
At $N=5\text{k}$ this quadratic term becomes the dominant computational overhead.\newline

\noindent\textbf{IConE-SIGreg and IConE-VCReg as efficient alternatives.}
The IConE-SIGreg and IConE-VCReg variants replace the diversity regularisation with losses that calculate statistical properties of the instance embeddings, eliminating the need to calculate the $N \times N$ matrix.
As shown in Figures~\ref{fig:supp_gpu_memory_sweep} and~\ref{fig:supp_wallclock_sweep}, both variants scale on par with the best baselines across all dataset sizes and batch sizes, while retaining the instance-centric inductive bias of the full IConE objective.
IConE-SIGreg and IConE-VCReg are therefore the recommended choices when training on larger datasets or under memory-constrained conditions.

\begin{table}[!t]
\centering
\caption{\textbf{Peak GPU memory scaling.} Algorithmic peak GPU memory (GB) for $224^2$ inputs at dataset size $N$ and embedding dimension $d{=}384$. Lowest per column in \textbf{bold}.}
\label{tab:icone_memory_scaling}
\scriptsize
\setlength{\tabcolsep}{2.5pt}
\begin{tabular}{@{}l ccc ccc ccc ccc@{}}
\toprule
& \multicolumn{3}{c}{$N{=}5$k} & \multicolumn{3}{c}{$N{=}10$k} & \multicolumn{3}{c}{$N{=}50$k} & \multicolumn{3}{c}{$N{=}100$k} \\
\cmidrule(lr){2-4} \cmidrule(lr){5-7} \cmidrule(lr){8-10} \cmidrule(lr){11-13}
\textbf{Method}
& $B{=}4$ & $B{=}16$ & $B{=}64$ & $B{=}4$ & $B{=}16$ & $B{=}64$ & $B{=}4$ & $B{=}16$ & $B{=}64$ & $B{=}4$ & $B{=}16$ & $B{=}64$ \\
\midrule
IConE        & 2.2 & 3.7 & 10.0 & 6.6 & 8.3 & 15.0 & 158.2 & 164.8 & 192.5 & 637.2 & 660.5 & 759.2 \\
IConE-VCReg  & \textbf{1.0} & \textbf{2.5} & \textbf{8.9} & \textbf{1.0} & \textbf{2.6} & \textbf{8.9} & \textbf{1.3} & \textbf{2.9} & \textbf{9.2} & \textbf{1.7} & \textbf{3.2} & \textbf{9.6} \\
IConE-SIGReg & 1.3 & 2.9 & 9.2 & 1.6 & 3.2 & 9.5 & 3.9 & 5.5 & 11.8 & 6.9 & 8.4 & 14.7 \\
\bottomrule
\end{tabular}
\end{table}

\section{Representation Quality Metrics}
\label{sec:appendix:metrics}

Beyond downstream task performance, we provide comprehensive measurements of representation geometry. These metrics offer insight into the intrinsic quality of learned representations independent of specific downstream tasks.

\subsection{Metric Definitions}

Let $\mathbf{Z} \in \mathbb{R}^{N \times d}$ denote the matrix of $N$ embeddings with dimension $d$, and let $\bar{\mathbf{Z}} = \mathbf{Z} - \frac{1}{N}\mathbf{1}\mathbf{1}^\top\mathbf{Z}$ be the centered embeddings.\newline

\noindent\textbf{Effective Rank.} The effective rank measures dimensional utilization via the entropy of normalized singular values. Let $\sigma_1, \ldots, \sigma_r$ be the singular values of $\bar{\mathbf{Z}}$, and define the normalized distribution $p_i = \sigma_i / \sum_j \sigma_j$. The effective rank is:
\begin{equation}
    \text{EffRank}(\mathbf{Z}) = \exp\left(-\sum_{i=1}^{r} p_i \log p_i\right).
\end{equation}
Values range from 1 (complete collapse to a single direction) to $\min(N, d)$ (full rank utilization).\newline

\noindent\textbf{RankMe.} RankMe is closely related to effective rank, but is commonly used in SSL as a spectrum-based proxy for representation quality. For the embedding matrix $\mathbf{Z}$ with singular values $\sigma_1, \ldots, \sigma_{\min(N,d)}$, define $p_k = \sigma_k / \|\sigma\|_1$:
\begin{equation}
    \text{RankMe}(\mathbf{Z}) = \exp\left(-\sum_{k} p_k \log p_k\right).
\end{equation}
Higher values indicate that variance is distributed across more dimensions rather than concentrated in a few, suggesting resistance to dimensional collapse.\newline

\noindent\textbf{LiDAR.} LiDAR (Linear Discriminant Analysis Rank) measures representation quality by computing the effective rank of the Linear Discriminant Analysis matrix associated with the SSL task. Using augmented views as surrogate classes, LiDAR computes:
\begin{align}
    \Sigma_b &= \mathbb{E}_{x}[(\mu_x - \mu)(\mu_x - \mu)^\top], \\
    \Sigma_w &= \mathbb{E}_{x}\mathbb{E}_{\tilde{x}|x}[(f(\tilde{x}) - \mu_x)(f(\tilde{x}) - \mu_x)^\top] + \delta I,
\end{align}
where $\mu_x$ is the mean embedding across augmentations of sample $x$, $\mu$ is the global mean, and $\delta$ is a small regularization constant. We construct an LDA-style surrogate-task matrix from between-instance and within-instance scatter and report its effective rank. Higher values indicate more discriminative directions in the representation space.\newline


\noindent\textbf{Alignment and Uniformity.} We measure two key properties of contrastive representations. Let $\tilde{\mathbf{z}} = \mathbf{z} / \|\mathbf{z}\|_2$ denote L2-normalized embeddings. Alignment measures the expected distance between positive pairs, defined as two augmented views $(x, x^+)$ of the same instance:
\begin{equation}
    \mathcal{L}_{\text{align}} = \mathbb{E}_{(x, x^+) \sim p_{\text{pos}}} \left[\|\tilde{\mathbf{z}}_x - \tilde{\mathbf{z}}_{x^+}\|_2^\alpha\right],
\end{equation}
where $\alpha = 2$. Lower values indicate that different augmentations of the same instance map to similar representations.

Uniformity measures how well the representations are spread on the unit hypersphere using the Gaussian potential kernel:
\begin{equation}
    \mathcal{L}_{\text{uniform}} = \log \mathbb{E}_{x, y \overset{\text{i.i.d.}}{\sim} p_{\text{data}}} \left[e^{-t\|\tilde{\mathbf{z}}_x - \tilde{\mathbf{z}}_y\|_2^2}\right],
\end{equation}
where $t = 2$ is the temperature parameter. More negative values indicate more uniform distribution on the hypersphere; values approaching 0 indicate collapse where all representations cluster together.

\subsection{Results Across Methods and Batch Sizes}

\begin{figure*}[!htbp]
\centering
\includegraphics[width=\textwidth]{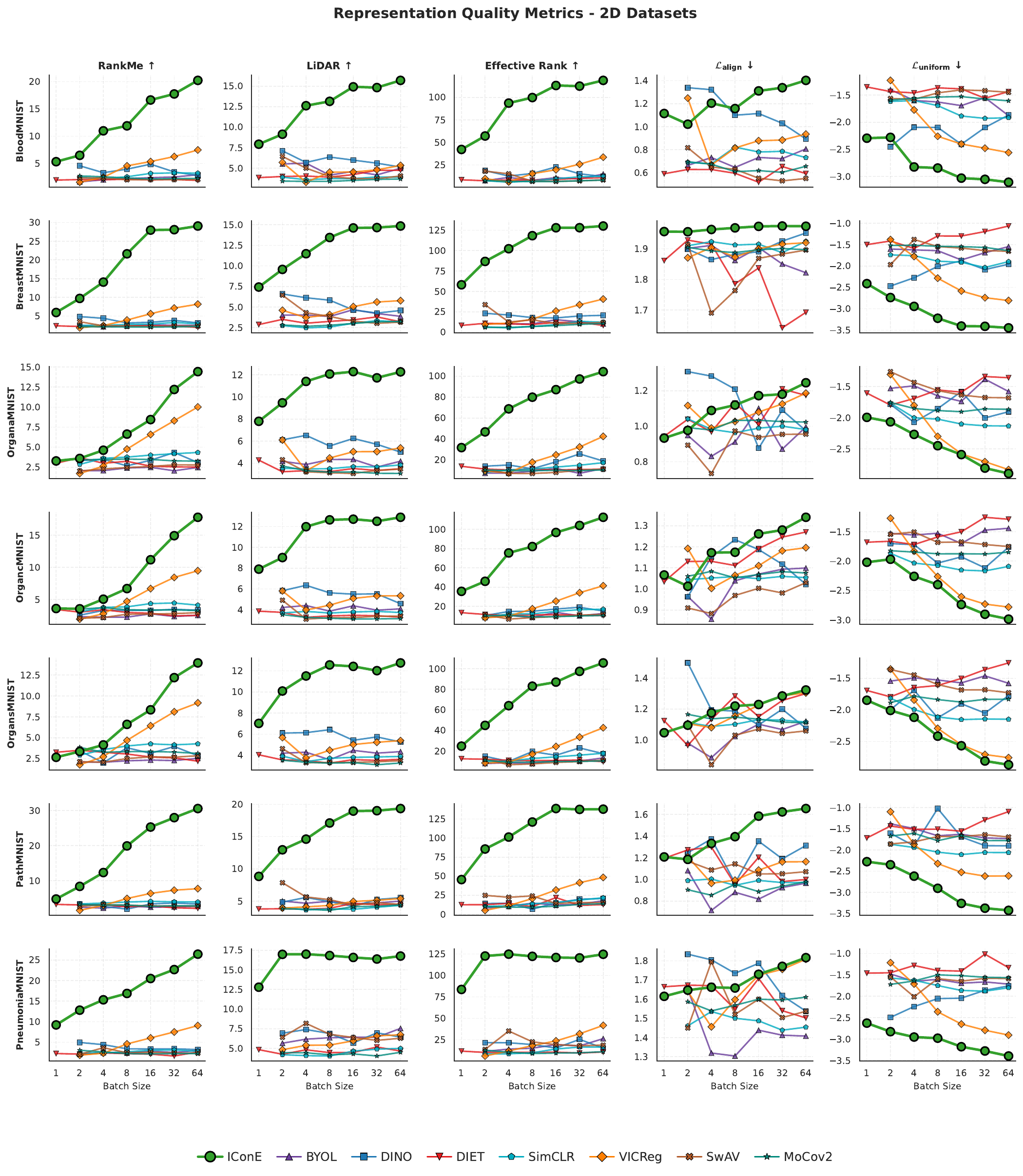}
\caption{\textbf{Representation quality metrics on 2D MedMNIST datasets.} Each row shows a different dataset; each column shows a different metric as a function of batch size. Higher is better for RankMe, LiDAR, and Effective Rank (indicating better dimensional utilization). Lower is better for $\mathcal{L}_{\text{align}}$ (tighter same-class clusters) and $\mathcal{L}_{\text{uniform}}$ (more uniform coverage). IConE (green) consistently achieves the best collapse-related metrics while maintaining competitive alignment and uniformity, especially at small batch sizes where baseline methods exhibit severe collapse.}
\label{fig:repr_metrics_2d}
\end{figure*}

\begin{figure*}[!htbp]
\centering
\includegraphics[width=\textwidth]{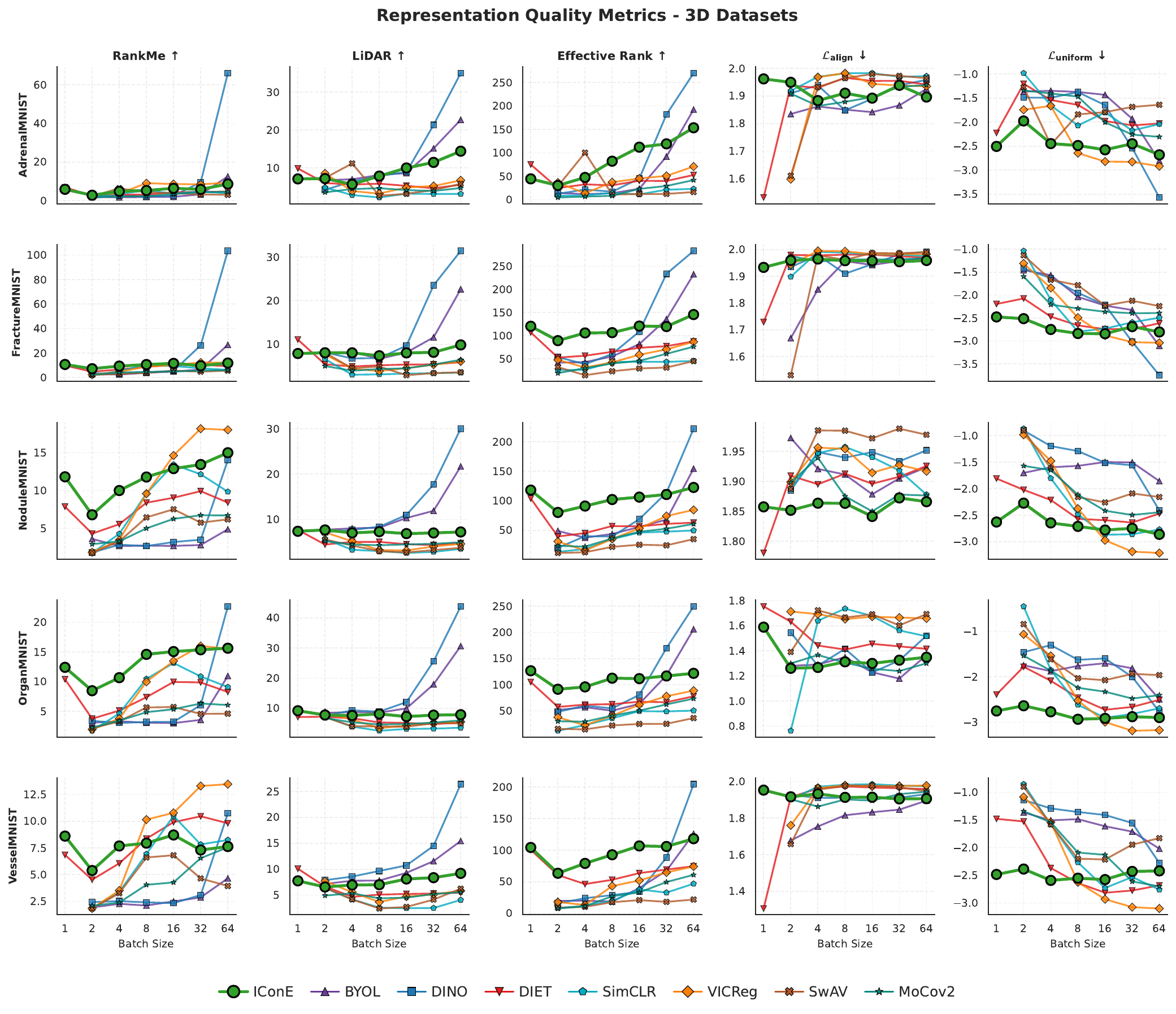}
\caption{\textbf{Representation quality metrics on 3D MedMNIST datasets.} Same format as \cref{fig:repr_metrics_2d}. IConE's advantages are even more pronounced in 3D, where the increased computational cost of volumetric data makes small-batch training essential. At $B=1$ and $B=2$, baseline methods show near-complete dimensional collapse (low RankMe, LiDAR, Effective Rank), while IConE maintains high-quality representations with good dimensional utilization.}
\label{fig:repr_metrics_3d}
\end{figure*}

\cref{fig:repr_metrics_2d} and \cref{fig:repr_metrics_3d} present representation quality metrics for all methods across batch sizes on 2D and 3D datasets respectively. All embeddings are standardized (zero mean, unit variance per dimension) before computing metrics to ensure fair comparison across methods that may encode information at different scales.

IConE consistently achieves superior scores on collapse-related metrics (RankMe, LiDAR, Effective Rank) while maintaining competitive alignment and uniformity, confirming that our explicit orthogonality regularization successfully prevents dimensional collapse without sacrificing representation quality.

\section{Natural Image Evaluation}
\label{sec:natural_image}

\begin{figure}[!t]
    \centering
    \includegraphics[width=\linewidth]{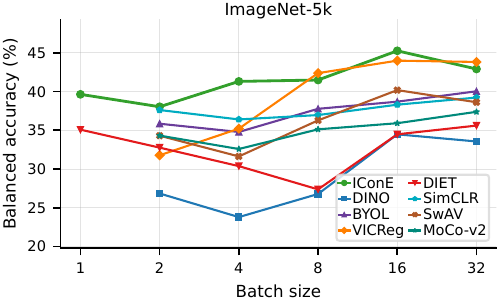}
    \caption{\textbf{Natural image evaluation.} Linear-probe top-1 balanced accuracy across batch sizes on a 20-class ImageNet subset (250 images per class). IConE retains its batch-size stability outside the biomedical domain.}
    \label{fig:imagenet_subset}
\end{figure}

To assess whether IConE's advantages hold outside the biomedical domain while remaining in our target small-data, small-batch regime, we evaluate on a subset of ImageNet. We randomly select 20 classes and subsample 250 images per class ($N{=}5$k total), and follow the same pretraining and linear-probe protocol as our main experiments. As shown in \cref{fig:imagenet_subset}, the same pattern emerges: IConE remains stable across batch sizes while batch-dependent baselines degrade as the batch shrinks, indicating that the method's batch-size robustness is not specific to medical imaging.

\section{Synthetic 2D Experiment}
\label{sec:toy_experiment}

To provide intuition for how IConE learns representations and to demonstrate the necessity of each loss component, we conduct an ablation study on a controlled synthetic dataset where we can directly visualize the learned representations.

\subsubsection{Experimental Setup}

\begin{figure}[!htbp]
\centering
\includegraphics[width=\textwidth]{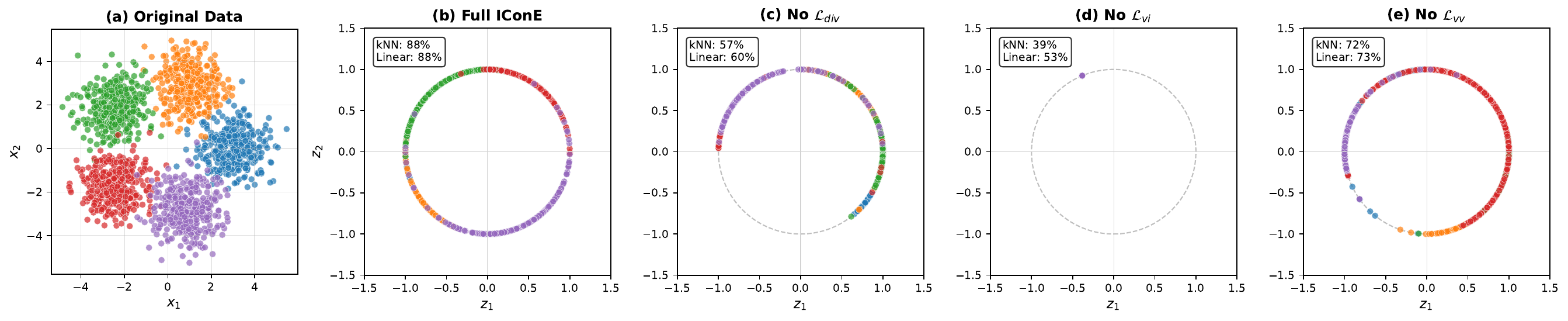}
\caption{\textbf{Final representations for each ablation variant.} (a) Original 2D Gaussian mixture data with 5 overlapping classes. (b) Full IConE achieves well-separated clusters on the unit circle (88\% kNN, 88\% linear). (c) Without $\mathcal{L}_{div}$, representations collapse into a small arc (57\% kNN). (d) Without $\mathcal{L}_{vi}$, no learning occurs (39\% kNN, near random). (e) Without $\mathcal{L}_{vv}$, separation is degraded (72\% kNN).}
\label{fig:toy_representations}
\end{figure}

\noindent\textbf{Synthetic Dataset.} We generate a 2D Gaussian Mixture Model (GMM) consisting of 5 classes with 350 instances per class (1,750 total samples). Class centers are placed uniformly on a circle of radius 3.0 at angles $\theta_k = \frac{2\pi k}{5}$ for $k \in \{0, 1, 2, 3, 4\}$. Each instance is sampled from a Gaussian distribution centered at its class center with standard deviation $\sigma_{\text{class}} = 0.8$, creating overlapping clusters that represent a classification task (~\cref{fig:toy_representations}a).\newline

\noindent\textbf{Augmentation Strategy.} For each instance, we generate 4 augmented views by adding Gaussian noise with $\sigma_{\text{aug}} = 0.15$ to the instance center. This simulates the effect of data augmentation in real SSL settings, where different views of the same image should map to similar representations.\newline

\noindent\textbf{Model Architecture.} We use a simple 2-layer MLP encoder (input $\rightarrow$ 64 hidden $\rightarrow$ 64 hidden $\rightarrow$ 2 output) mapping 2D inputs to 2D normalized representations on the unit circle. Using 2D representations allows direct visualization without dimensionality reduction artifacts. Instance embeddings are initialized with small random values ($\sigma = 0.02$) and L2-normalized.\newline

\noindent\textbf{Training.} All variants are trained for 300 epochs using Adam optimizer with learning rate $10^{-3}$ and batch size 128. We evaluate using 5-NN classification and linear probe accuracy on a 70/30 train/test split.

\noindent\textbf{Evaluation Metrics.} We compute:
\begin{itemize}
    \item \textbf{Alignment loss}: $\mathcal{L}_{\text{align}} = \mathbb{E}_{(x,y) \sim p_{\text{same}}} \left[ \|f(x) - f(y)\|^2 \right]$, measuring within-class compactness (lower is better). This is a class-conditional adaptation of the original metric.
    \item \textbf{Uniformity loss}: $\mathcal{L}_{\text{uniform}} = \log \mathbb{E}_{(x,y) \sim p_{\text{all}}} \left[ e^{-2\|f(x) - f(y)\|^2} \right]$, measuring feature space coverage (lower is better).
\end{itemize}

\subsubsection{Ablation Variants}

We train four variants to isolate the contribution of each loss component(see \cref{tab:ablation_variants}).

\begin{table}[h]
\centering
\caption{Ablation variants configuration.}
\label{tab:ablation_variants}
\begin{tabular}{lccc}
\toprule
Variant & $\mathcal{L}_{vv}$ & $\mathcal{L}_{vi}$ & $\mathcal{L}_{div}$ \\
\midrule
Full IConE & \checkmark & \checkmark & \checkmark \\
No $\mathcal{L}_{div}$ & \checkmark & \checkmark & $\times$ \\
No $\mathcal{L}_{vi}$ & \checkmark & $\times$ & \checkmark \\
No $\mathcal{L}_{vv}$ & $\times$ & \checkmark & \checkmark \\
\bottomrule
\end{tabular}
\end{table}

\subsubsection{Results and Analysis}

\noindent\textbf{Final Representations.} ~\cref{fig:toy_representations} shows the original data and the learned representations for each ablation variant. The full IConE model learns well-separated class clusters on the unit circle, achieving 88\% kNN and 88\% linear probe accuracy. Removing $\mathcal{L}_{div}$ causes some representation collapse, all points cluster into a half-circle arc, reducing linea probe accuracy to 60\%. Without $\mathcal{L}_{vi}$, no meaningful learning occurs (39\% kNN accuracy, near random for 5 classes). Removing $\mathcal{L}_{vv}$ still allows learning but with degraded performance (73\% linear probe accuracy).\newline

\begin{figure}[!htbp]
\centering
\includegraphics[width=\textwidth]{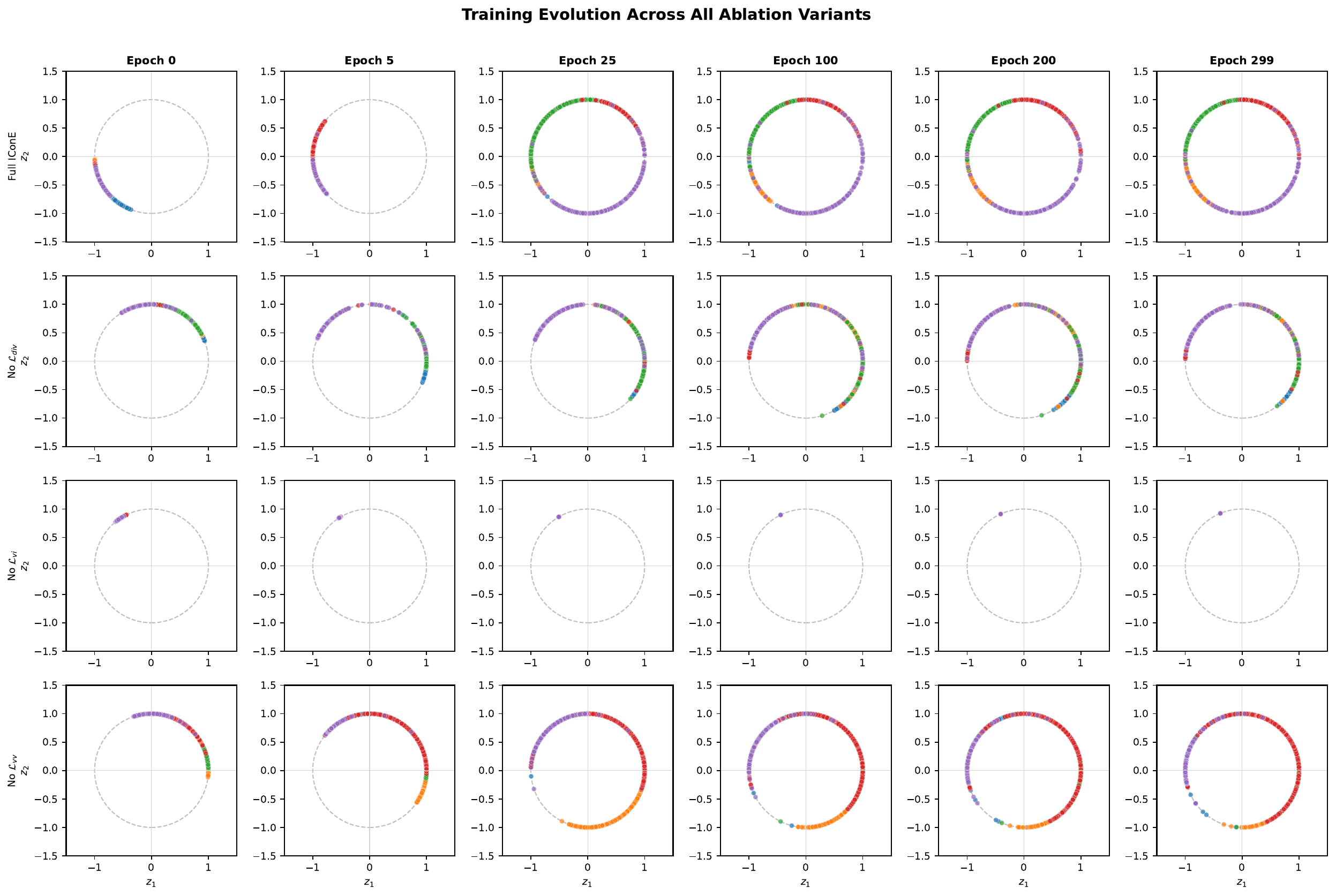}
\caption{\textbf{Training evolution across all ablation variants.} Each row shows a different variant; each column shows representations at a specific epoch (0, 5, 25, 100, 200, 299). Points are colored by class label.}
\label{fig:toy_evolution}
\end{figure}

\noindent\textbf{Training Dynamics.} ~\cref{fig:toy_evolution} shows how representations evolve during training for each ablation variant across epochs.\newline

\begin{figure}[!htbp]
\centering
\includegraphics[width=\textwidth]{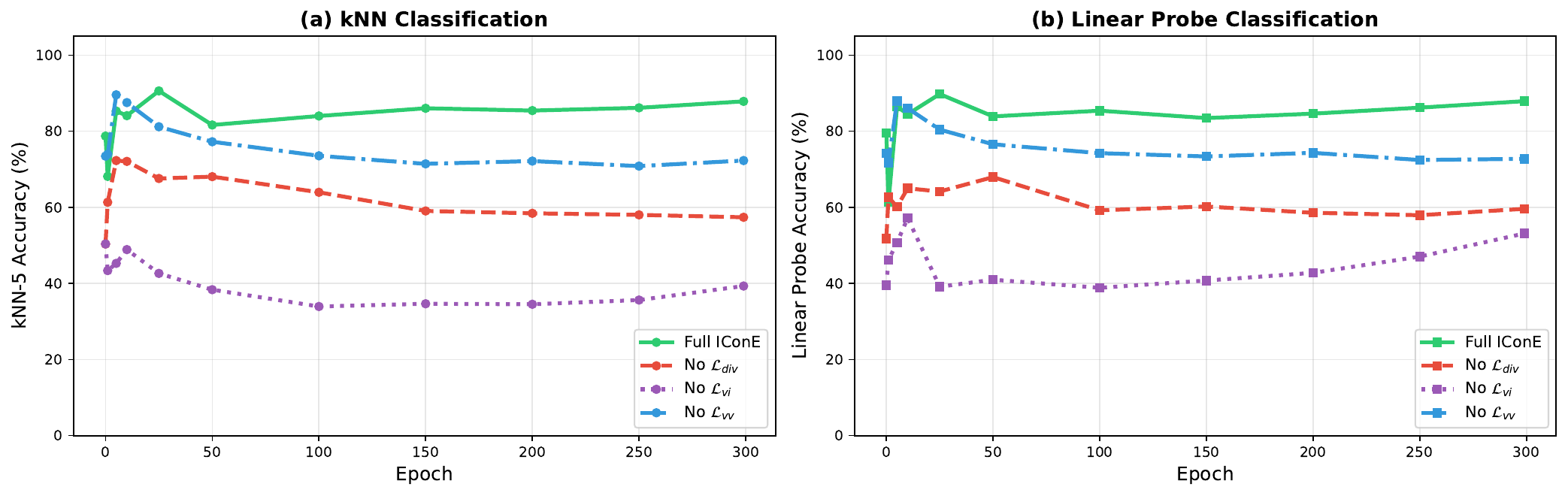}
\caption{\textbf{Classification accuracy during training.} (a) kNN-5 accuracy. (b) Linear probe accuracy. Full IConE maintains stable high performance, while ablated variants show degraded or unstable learning.}
\label{fig:toy_accuracy}
\end{figure}

\noindent\textbf{Downstream Performance.} ~\cref{fig:toy_accuracy} presents the kNN-5 and linear probe classification accuracy throughout training for all variants.\newline

\begin{figure}[!htbp]
\centering
\includegraphics[width=\textwidth]{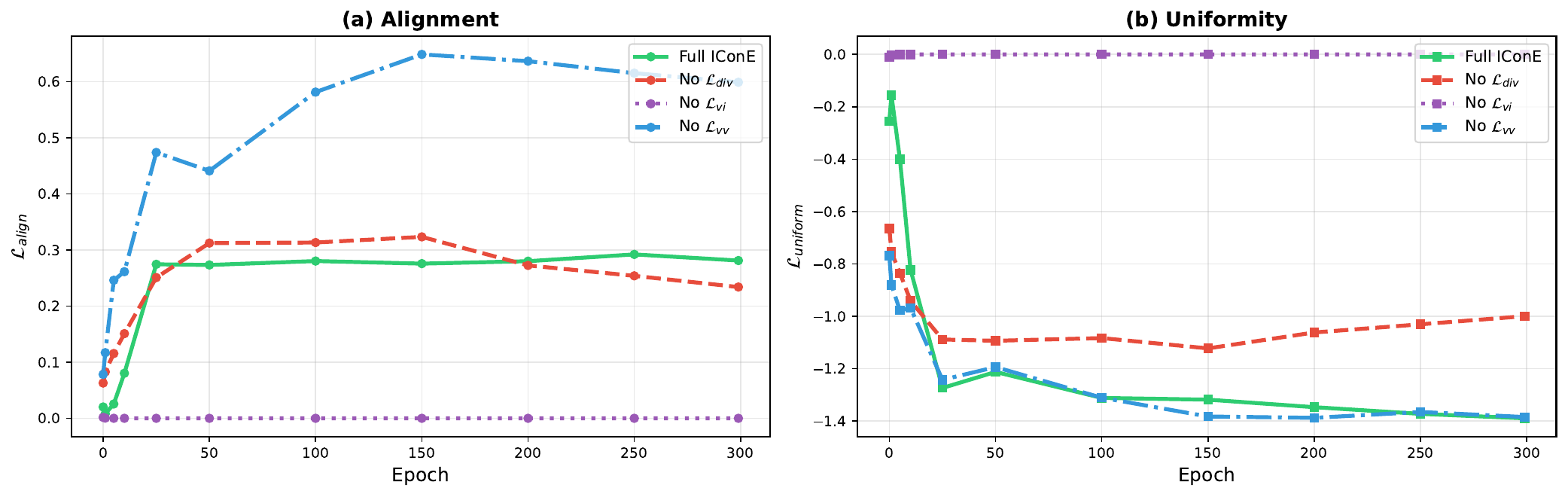}
\caption{\textbf{Alignment and uniformity metrics during training.} (a) Alignment loss (lower = tighter same-class clusters). (b) Uniformity loss (lower = more spread). Full IConE achieves the best trade-off with both low alignment and low uniformity. Without $\mathcal{L}_{vv}$, alignment suffers; without $\mathcal{L}_{div}$, uniformity suffers.}
\label{fig:toy_alignment_uniformity}
\end{figure}

\noindent\textbf{Alignment-Uniformity Trade-off.} ~\cref{fig:toy_alignment_uniformity} shows the alignment and uniformity metrics during training. This analysis reveals the distinct roles of each loss component:

\begin{itemize}
    \item \textbf{$\mathcal{L}_{div}$ controls uniformity}: Without the diversity loss, uniformity remains poor ($\mathcal{L}_{\text{uniform}} \approx -1.0$) compared to Full IConE ($\mathcal{L}_{\text{uniform}} \approx -1.4$). The diversity loss spreads instance embeddings across the hypersphere, and the encoder inherits this uniform distribution.
    
    \item \textbf{$\mathcal{L}_{vv}$ controls alignment}: Without view-view consistency, alignment degrades significantly ($\mathcal{L}_{\text{align}} \approx 0.60$) compared to Full IConE ($\mathcal{L}_{\text{align}} \approx 0.28$). The view-view loss ensures that different augmentations of the same instance produce consistent representations, which indirectly encourages same-class samples to cluster together.
    
    \item \textbf{$\mathcal{L}_{vi}$ is essential}: Without the view-instance alignment, both metrics remain at degenerate values ($\mathcal{L}_{\text{align}} \approx 0$, $\mathcal{L}_{\text{uniform}} \approx 0$), indicating no meaningful learning. This loss provides the fundamental learning signal that connects the encoder to the instance embeddings.
\end{itemize}

\subsubsection{Summary}

~\cref{tab:toy_results} summarizes the final performance of each variant. Full IConE achieves the best trade-off across most metrics, demonstrating that all three loss components are necessary for optimal representation learning.

\begin{table}[h]
\centering
\caption{Ablation study results on synthetic 2D data. Best values are shown in \textbf{bold}, second best are \underline{underlined}. $\dagger$ indicates degenerate values due to representation collapse.}
\label{tab:toy_results}
\begin{tabular}{lccccc}
\toprule
Variant & kNN-5 (\%)$\uparrow$ & Linear (\%)$\uparrow$ & $\mathcal{L}_{\text{align}}$$\downarrow$ & $\mathcal{L}_{\text{uniform}}$$\downarrow$ & Silhouette$\uparrow$ \\
\midrule
Full IConE & \textbf{87.9} & \textbf{87.9} & \underline{0.281} & \textbf{-1.389} & \textbf{0.475} \\
No $\mathcal{L}_{div}$ & 57.4 & 59.6 & \textbf{0.234} & -0.999 & 0.071 \\
No $\mathcal{L}_{vi}$ & 39.3 & 53.1 & 0.000$^\dagger$ & -0.000$^\dagger$ & 0.015 \\
No $\mathcal{L}_{vv}$ & \underline{72.3} & \underline{72.8} & 0.599 & \underline{-1.384} & \underline{0.255} \\
\bottomrule
\end{tabular}
\end{table}

This toy experiment demonstrates that IConE's three loss components serve complementary roles: $\mathcal{L}_{vi}$ provides the core instance discrimination signal, $\mathcal{L}_{div}$ prevents representation collapse by enforcing uniformity, and $\mathcal{L}_{vv}$ improves alignment by enforcing augmentation consistency. The optimal representation quality requires all three components working together to achieve both low alignment (compact same-class clusters) and low uniformity (uniform coverage of the representation space).


\end{document}